\documentclass[10pt,twocolumn,letterpaper]{article}

\usepackage[pagenumbers]{cvpr} % To force page numbers, e.g. for an arXiv version

\definecolor{cvprblue}{rgb}{0.21,0.49,0.74}
\usepackage[pagebackref,breaklinks,colorlinks,allcolors=cvprblue]{hyperref}
\usepackage{tabularx}
\usepackage{ragged2e}
\usepackage{booktabs}
\usepackage{adjustbox}
\usepackage{multirow}
\usepackage{graphicx}
\usepackage[table]{xcolor}
\usepackage{csquotes}
\usepackage{amsmath,amssymb,amsthm,mathtools}
\usepackage{makecell}
\usepackage{comment}

\def\confName{CVPR}
\def\confYear{2027}

\AtEndPreamble{
  \crefname{appendix}{App.}{Apps.}
  \Crefname{appendix}{App.}{Apps.}
  \crefname{subappendix}{App.}{Apps.}
  \Crefname{subappendix}{App.}{Apps.}
}

\newcolumntype{C}{>{\centering\arraybackslash}p{0.62cm}}

\newcommand{\mypara}[1]{\smallskip\noindent\textbf{#1.}}

\newcommand{\UB}[1]{\textcolor{gray}{#1}}

\newcommand{\calib}{$^\dagger$}
\newcommand{\aq}[1]{%
  \raisebox{0.08ex}{\scalebox{0.65}[0.75]{$<$}}#1%
  \raisebox{0.08ex}{\scalebox{0.65}[0.75]{$>$}}%
}

\DeclareMathOperator{\tr}{tr}
\providecommand{\UB}[1]{\textcolor{gray}{#1}}
\providecommand{\calib}{$^\dagger$}

\title{Rescaling OLS-Merged Adapters To Improve Personalized Image Generation}

\author{Leander Girrbach\textsuperscript{1,2} \qquad  Niklas Schweiger\textsuperscript{1} \qquad Zeynep Akata\textsuperscript{1,2} \\
\textsuperscript{1}Technical University of Munich (TUM),\quad Munich Center for Machine Learning (MCML) \\ \textsuperscript{2} Helmholtz Munich
}

\begin{document}
\maketitle

\begin{abstract}
Merging multiple adapters into a single image generation model reduces deployment costs while preserving diverse task capabilities or personalization modes. State-of-the-art methods such as RegMean, Mix-of-Show, and Subspace Routing fit the outputs of the merged model to those of the individual adapters using ordinary least squares (OLS). We show that this reconstruction objective systematically shrinks task-specific signals whenever the activation distributions of different tasks overlap, producing a merged adapter that is too weak for downstream use. We derive a decomposition that quantifies this shrinkage and propose rescaling the merged update to restore signal strength. On multi-subject personalization with up to 30 merged adapters, rescaling substantially improves subject fidelity over unscaled baselines. Our approach also improves results on style transfer, concept erasure, image classification, and LLM-based text classification.
\end{abstract}

\section{Introduction}
\label{sec:intro}

\suppressfloats[t]
\begin{figure}[t]
\centering\includegraphics[width=\linewidth]{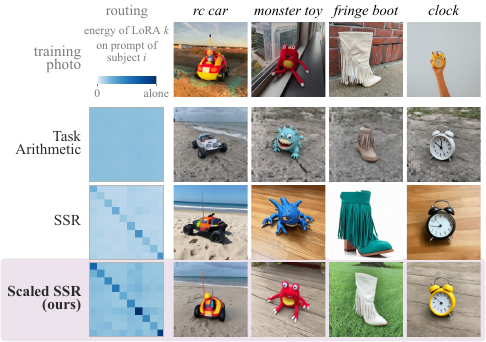}
\caption{30 DreamBooth LoRAs merged into SD3.5-Medium. Heatmaps: energy of LoRA $k$ on subject $i$, relative to LoRA $k$ alone. Task Arithmetic spreads energy uniformly, SSR routes correctly but too weakly, and Scaled SSR (ours) restores subjects.}\label{fig:teaser}
\end{figure}

Low-rank adapters (LoRAs) \cite{hu2022lora} have become the standard way to personalize text-to-image diffusion models. A few images are enough to teach a model a new subject \cite{ruiz2023dreambooth}, an artistic style, or an erasure that removes a concept from the model \cite{lu2024mace,gandikota2024unified}. 
Many applications need several of these adapters in one model, for example to generate any of a set of personalized subjects or styles, or to erase a list of unwanted concepts at once.

One option is to train a single adapter on the images of all subjects together. This requires the training images of every subject, which are often unavailable because adapters are typically shared as weights without their data, and the training has to be repeated whenever a subject is added. 
The alternative is to combine independently trained adapters into one set of weights without further training, which is the model merging problem \cite{ilharco2023editing,yadav2023ties,yu2024dare,marczak2025isoc,stoica2025model,gu2023mix,shah2024ziplora}.

Merging becomes difficult when many adapters are combined. All adapters modify the same layers, and their prompts share most of their text embedding, so each adapter also changes the images generated for the other adapters' prompts. This interference grows with the number of adapters. 
Weight-space methods ignore the input distribution adapters were trained on, and their subject fidelity drops substantially as more adapters are merged (\cref{tab:dreambooth}).
Data-based methods instead fit the merged weights so that each merged layer reproduces the output of every adapter on that adapter's own inputs, which can be formalized as an ordinary least squares (OLS) problem \cite{jin2022dataless,gu2023mix,chen2025iteris,wei2026ssrmerge}.
Among them, Subspace Routing (SSR) \cite{wei2026ssrmerge} is particularly well suited for many adapters. It solves the OLS problem in the low-rank bottleneck of the LoRAs, so its closed-form solution scales with the total adapter rank instead of the layer width. SSR routes each prompt to the correct adapter, yet the subjects still fade as more adapters are merged (\cref{fig:teaser}).

\begin{figure*}[t]
\centering\includegraphics[width=\textwidth]{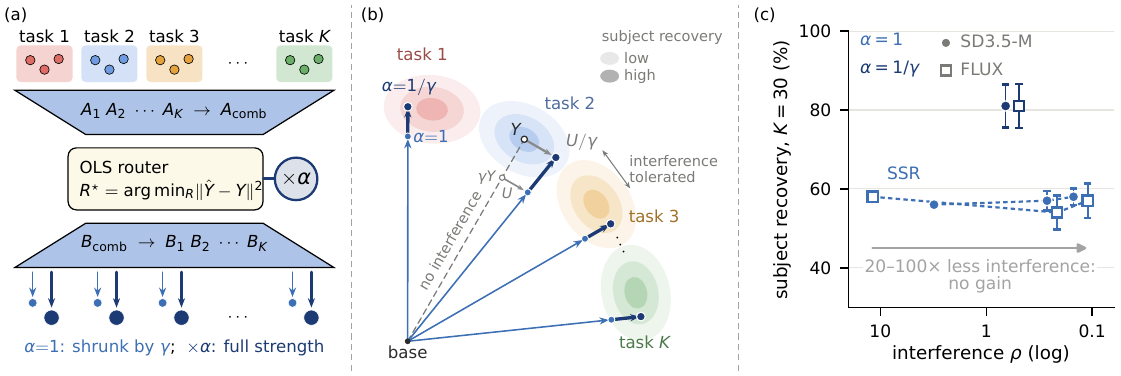}
\caption{Why OLS merging shrinks and how rescaling helps. \textbf{(a)} SSR merges $K$ LoRAs with an OLS router; we rescale the merged output by one factor $\alpha$. \textbf{(b)} Because the adapters' inputs overlap, OLS returns $\gamma Y + U$ instead of the target $Y$: the signal shrunk by $\gamma<1$, plus interference $U$. Rescaling by $1/\gamma$ restores the signal, and the task tolerates the larger interference $U/\gamma$. 
\textbf{(c)} With 30 LoRAs, cutting interference 20--100$\times$ leaves subject recovery (share of the base-to-single-LoRA DINO gap) at 54--58\,\%, while rescaling raises it to 81\,\%.}
\label{fig:concept}
\vspace{-10pt}
\end{figure*}

We trace this to a systematic weakness of OLS-based merging, which we call shrinkage. When the inputs of the adapters overlap, no merged layer can reproduce the output of every adapter exactly, and the least-squares solution returns a scaled-down version $\gamma Y$, with $\gamma<1$, of each adapter's output $Y$ plus interference from the other adapters (\cref{fig:concept}). This is related to regression dilution in statistics \cite{frost2000}. OLS treats a weakened signal and added interference as equally bad, but the generated images are more sensitive to the weakened signal: with 30 LoRAs, reducing the interference 20-100$\times$ doesn't help much, while undoing the shrinkage restores the subjects (\cref{fig:concept}\,(c)). Scaling the merged adapter up by a single factor therefore improves the merged model substantially.

In summary, our contributions are: 
\textbf{(1)} We show that OLS-based merging methods like SSR, RegMean, and Mix-of-Show shrink the output of adapters, and that the merged model is more sensitive to this weakened output than to interference from other adapters.
\textbf{(2)} From this analysis, we derive a rescaling factor $1/\gamma$ that is computed in closed form from the calibration data at almost no extra cost. Alternatively, the factor can be selected by a line search on validation data.
\textbf{(3)} By correcting shrinkage, Scaled SSR outperforms state-of-the-art merging methods when merging 30 DreamBooth subjects, 20 artistic styles, and concept erasers. Rescaling also improves RegMean and Mix-of-Show, and transfers to image and text classification.

\section{Related Work}
\label{sec:related-work}

{\noindent\textbf{Model Merging.}}
This paper is about improving techniques to merge multiple LoRA adapters into a single adapter. Model merging \cite{izmailov2018averaging,frankle2020linear,wortsman2022model,ilharco2023editing} in general is a broad topic, so we refer to surveys like \cite{li2025deep,yang2026model,he2025mergebench,hitit2026a} for an overview. Classical merging methods operate on task vectors, i.e., the weight delta between base and fine-tuned model variants \cite{ilharco2023editing}, and modify the task vectors before merging \cite{yadav2023ties,yu2024dare,deep2024della,wang2024localizing,davari2024model,huang2024emr}. Other methods decompose weight updates into complementary directions \cite{marczak2025isoc,zhang2026dcmerge,gu2026beyond,cho2026not,chaves2026bridging,wu2026gram,guo2026stay,sui2026coram,qiu2025superpose,wang2026orthogonal,shi2026pact,skorobogat2025subspace}. Some methods also use data to find relevant task subspaces, reduce noise, or mitigate task interference have been proposed \cite{matena2022merging,singh2020model,jin2022dataless,an2026mitigating,li2026model,cho2026error,zhang2025unraveling,wei2026ssrmerge,stoica2024zipit,xu2026acemerging,chen2025iteris}. For example, \citeauthor{jin2022dataless} (RegMean) and \citeauthor{wei2026ssrmerge} (SSR) propose approaches for merging weights to minimize the prediction difference between the merged model and the individual models. Data can also be used for direct training or gradient-free optimization of merged weights \cite{yang2024adamerging,yang2024representation,huang2024lorahub,akiba2025evolutionary,cao2026evolutionary}. Finally, LoRA adapters allow specifically tailored methods that operate on the individual projections and align shared directions of adapters \cite{stoica2025model,panariello2025accurate,zeng2025robustmerge,zheng2026decouple,ryum2026ct,amballa2026not,tang2026crowded,liu2026pgsb,lee2026labelfree,wei2026ssrmerge,cao2026evolutionary,zhang2025unraveling,yin2026chainlora,liang2025thanora}.

\mypara{Subject-Driven Image Generation}
Our main application is few-shot subject-driven image generation \cite{ruiz2023dreambooth}. Here, we adapt a text-to-image diffusion model \cite{ho2020denoising,nichol2022glide,rombach2022highresolution} to generate images with a specific subject in different contexts, using only a few images of the subject. Since this task is very relevant and popular, developing specific merging methods for this task is an active field of research \cite{shah2024ziplora,chen2025np,wei2026ssrmerge,gu2023mix,simsar2025loraclr,po2024orthogonal,ouyang2025k,yang2025lora,meral2025contrastive,wu2024mixture,zhong2024multilora}.

\mypara{Shrinkage and LoRA Scaling}
Our core insight is that merging methods that use ordinary least squares (OLS) systematically shrink the task signal, which is critical for the model. Shrinkage in OLS is a known phenomenon \cite{copas1983regression,copas1997using,wiener1949extrapolation,frost2000}, and our contribution is to analyze it for merging methods, show its relevance, and present a remedy through rescaling. PICO \cite{tang2026crowded} and LiNeS \cite{wang2025lines} propose different rescaling methods with very different motivations. In \cref*{sec:supp:ablations:picolines} we show that our method performs better.

\section{Theoretical Analysis of OLS-Based Merging}
\label{sec:theory}

{\noindent {\textbf{Activation-Based Merging as Least Squares}}}
Let $W_0$ denote a linear layer of the pretrained model and let $\{\Delta W_k\}_{k=1}^K$ be $K$ task-specific updates of this layer (e.g., LoRA adapters $\Delta W_k = B_kA_k$ \cite{hu2022lora} or full fine-tuning deltas). Given unlabeled calibration inputs $X_k \in \mathbb{R}^{d\times m}$ for each task, the output of update $k$ in isolation is $Y_k = \Delta W_k X_k$. Instead of averaging the -- possibly modified -- updates as in Task Arithmetic \cite{ilharco2023editing}, TIES \cite{yadav2023ties}, or DARE \cite{yu2024language}, a family of methods \cite{jin2022dataless,gu2023mix,wei2026ssrmerge} optimizes the merged update $\Delta W(\theta)$, parameterized linearly by $\theta$, to reproduce the standalone behavior of every task on its inputs:
\begin{equation}\label{eq:theory:ols}
    \theta^{\star} = \arg\min_{\theta} \sum_{k=1}^{N}
    \left\| \Delta W(\theta)\, X_k - Y_k \right\|_F^2 .
\end{equation}
Because \cref{eq:theory:ols} is quadratic in $\theta$, it is an Ordinary Least Squares (OLS) problem whose global minimizer depends on the calibration data only through activation statistics such as $\sum_k X_kX_k^\top$. Concrete instances differ in the parameterization: RegMean \cite{jin2022dataless} and the gradient fusion of Mix-of-Show \cite{gu2023mix} merge full weights, i.e., $\Delta W(\theta) = \theta$, solved in closed form or with LBFGS, respectively. SSR-Merge \cite{wei2026ssrmerge}, which we build on to scale to the largest number of merged LoRAs, stacks the adapters into $A_{\text{comb}} = [A_1^\top \cdots A_K^\top]^\top$ and $B_{\text{comb}} = [B_1 \cdots B_K]$ and learns a router $\theta = R$ in the low-rank bottleneck, $\Delta W(R) = B_{\text{comb}} R A_{\text{comb}}$, which keeps the merged model a LoRA and suppresses inteference between $A_k$ and unrelated $B_{k'}$ (see \cref*{sec:supp:detailederivation}). We analyze the structure of \cref{eq:theory:ols} that applies to all members of this family. In the following, we write $\hat Y_k = \Delta W(\theta^\star) X_k$ for the merged prediction.

\mypara{OLS Suffers from Shrinkage}
In reality, no single linear update can perfectly disentangle all $K$ tasks simultaneously. This leads to a problem called shrinkage: to prevent off-target interference from inflating the loss in \cref{eq:theory:ols}, OLS systematically shrinks the on-target signal toward zero.
To formalize this, we decompose the predictions $\{\hat{Y}_k\}_{k=1}^K$ into a component that is collinear with the true targets $\{Y_k\}_{k=1}^K$ and an orthogonal residual $\{U_k\}_{k=1}^K$ that captures the off-target interference:
\begin{equation}\label{eq:theory:shrinkage:yhat}
    \hat{Y}_k = \gamma Y_k + U_k, \quad \text{such that} \quad \sum_{k=1}^{K} \langle U_k, Y_k \rangle_F = 0,
\end{equation}
where $\langle A, B \rangle_F = \mathrm{tr}(A^\top B)$ is the Frobenius inner product. By orthogonal projection, the factor $\gamma$ is
\begin{equation}\label{eq:theory:shrinkage:gamma}
    \gamma = \frac{\sum_{k=1}^{K} \langle \hat{Y}_k, Y_k \rangle_F}{\sum_{k=1}^{K} \|Y_k\|_F^2}.
\end{equation}
For example, $\gamma = 0.6$ means that, on average across tasks, the merged update preserves only $60\%$ of the target adapter size along the true task directions. Similarly, we define
\begin{equation}\label{eq:theory:shrinkage:he}
    \hat{h} = \frac{\sum_{k=1}^{K} \|\hat{Y}_k\|_F^2}{\sum_{k=1}^{K} \|Y_k\|_F^2}, \qquad
    e = \frac{\sum_{k=1}^{K} \|U_k\|_F^2}{\sum_{k=1}^{K} \|Y_k\|_F^2}.
\end{equation}
$\hat{h}$ captures the ratio of the total size of the router outputs compared with the size of the targets, and $e$ captures the size of the off-target interference compared to the size of the targets.
Because $\sum_{k=1}^K \langle U_k, Y_k \rangle_F = 0$, we have
\begin{equation}\label{eq:theory:shrinkage:pythagoras}
    \hat{h} = \gamma^2 + e,
\end{equation}
and can conclude why \eqref{eq:theory:ols} gives $\gamma < 1$, i.e, shrinkage:
Suppose we scale $\theta^\star$ by factor $c\in \mathbb{R}$. Because $\Delta W(\theta)$ is linear in $\theta$ and $\theta^\star$ is the global minimizer, $c = 1$ minimizes
\begin{align}
    \phi(c) 
    &= \sum_{k=1}^{K} \| c \hat{Y}_k - Y_k \|_F^2 \\
    &= \left( c^2 \hat{h} - 2 c \gamma + 1 \right) \sum_{k=1}^{K} \|Y_k\|_F^2.
\end{align}
Setting the derivative $\phi'(1) = 0$ gives 
\begin{equation}
    \sum_{k=1}^{K} \langle \hat{Y}_k - Y_k, \hat{Y}_k \rangle_F = 0,
\end{equation}
which implies $\hat{h} = \gamma$. 
Substituting $\hat{h} = \gamma$ into \cref{eq:theory:shrinkage:pythagoras} gives $\gamma = \gamma^2 + e$, or equivalently,
\begin{equation}\label{eq:theory:shrinkage:gamma_bound}
    \gamma = 1 - \frac{e}{\gamma}, \qquad e = \gamma(1 - \gamma).
\end{equation}
Since $e > 0$ whenever task subspaces overlap, i.e., in practice, $\gamma < 1$. Intuitively, if OLS were to output full signal strength ($\gamma = 1$) in the presence of orthogonal error ($e > 0$), its total size would be $\hat{h} = 1 + e > 1$, which increases the Euclidean distance to $Y_k$. OLS reduces this Euclidean error by shrinking both the signal and the interference by $\gamma < 1$.

\mypara{Local Reconstruction vs.\ Task Loss}
While $\theta^{\star}$ is optimal for reconstructing the local target $Y_k$, this is not identical with optimality for the global task loss, such as classification accuracy or image generation quality. Here, we observe (empirically and intuitively) that the sensitivity of the global loss $\mathcal{L}$ to shrinkage of $Y_k$ differs from that to shrinking the off-target interference, which is generally noise that can cancel across layers or is mitigated by the model's general robustness. In contrast, the task-specific updates $\Delta W_k$ were specifically trained to inject the full target direction $Y_k$ and not $\gamma Y_k$.

Specifically, assume that the global loss $\mathcal{L}$ is minimal for a factor $s^\star$ of $Y_k$. Since we assume the updates $\Delta W_k$ are well-trained, generally $s^{\star} \approx 1$. An intuitive solution to address shrinkage is to scale the merged update $\Delta W(\theta^{\star})$ by a factor $\alpha$ (for SSR-Merge, this amounts to scaling the router $R^{\star}$) that restores the factor $s^{\star}$ of $Y_k$ without over-amplifying interference. Denote $a$ the curvature $\frac{\partial^2\mathcal{L}}{\partial s^2}$ of $s$ towards $s^\star$, and $b$ the curvature of scaling the off-target interference $U_k$, i.e., how sensitive the global loss is to noise.

To find the optimal value of $\alpha^\star$, we expand the global loss around $\alpha$ as the second-order Taylor expansion
\begin{equation}\label{eq:theory:downstream:quad}
    \mathcal{L}(\alpha) \approx \mathcal{L}_{\min} + \frac{a}{2}(\alpha\gamma - s^\star)^2 + \frac{b}{2}\alpha^2e,
\end{equation}
where $\mathcal{L}_{\min}$ is the loss with optimal factor $s^\star$.
Differentiating \cref{eq:theory:downstream:quad} with respect to $\alpha$ and setting $\mathcal{L}'(\alpha^\star) = 0$ yields the optimum
\begin{equation}\label{eq:theory:downstream:alpha_star}
    \alpha^\star = \frac{s^\star \gamma}{\gamma^2 + \frac{eb}{a}} = \frac{s^\star}{\gamma} \cdot \frac{1}{1 +  \frac{eb}{a\gamma^2}}.
\end{equation}
\Cref{eq:theory:downstream:alpha_star} disentangles the two components that influence $\alpha^\star$: (i) The first factor reverts the shrinkage $\gamma < 1$ to restore the target signal amplitude $s^\star$ expected by the global loss. (ii) The second factor prevents $\alpha$ from over-amplifying the off-target interferences $U_k$.
When off-target interference $e$ is small and $s^\star \approx 1$, the second factor is close to $1$, motivating $\hat{\alpha} = 1/\gamma$ as heuristic. Conversely, when $e$ is large, $1 + \frac{be}{a\gamma^2}$ pulls $\alpha^\star$ back toward $1$.

\mypara{Finding the Rescaling Factor}
Our analysis in \cref{sec:theory} provides a method to improve any OLS-based merge by scaling the merged update $\Delta W(\theta^\star)$ by $\alpha$ , and gives a closed-form solution for the optimal factor $\alpha^\star$. However, in practice several quantities in \cref{eq:theory:downstream:alpha_star} cannot be estimated empirically: $s^{\star}$ is an oracle of the optimal $Y_k$ scaling which is generally inaccessible, and $\frac{b}{a}$ depends on second-order gradients which are costly to estimate. This leaves two viable options: Use the $1/\gamma$ heuristic proposed above, or, assume a single common factor $\alpha$ for the whole model and conduct a line-search on a validation set to find the best value. In our experiments, we evaluate both variants, but also use a common value $\gamma$ across layer groups (image stream, text stream, and shared stream; three values) to stabilize the heuristic, see \cref*{sec:supp:perlayer} for a comparison. We compare the scaling factors returned by the $1/\gamma$ heuristic and line search in \cref*{sec:supp:gammaalpha}. In our experiments, we generally apply scaling to SSR \cite{wei2026ssrmerge}, because it is the most computationally efficient OLS method for LoRA merging. Experiments on RegMean and Mix-of-Show are in \cref{sec:ablations:othermethods}. Technical details on how we compute $\gamma$ for the different methods are in \cref*{sec:supp:gamma}.

\section{Experimental Evaluation}
\label{sec:experiments}

\begin{table*}[t]
\centering
%\footnotesize
\setlength{\tabcolsep}{3pt}
%\resizebox{\textwidth}{!}{%
\begin{tabular}{l cccccccc cccccccc}
\toprule
\multirow{3.5}{*}{\textbf{Method}} & \multicolumn{8}{c}{\textbf{FLUX.1-dev}} & \multicolumn{8}{c}{\textbf{SD3.5-Medium}} \\
\cmidrule(lr){2-9} \cmidrule(lr){10-17}
& \multicolumn{2}{c}{$9$} & \multicolumn{2}{c}{$15$} & \multicolumn{2}{c}{$20$} & \multicolumn{2}{c}{$30$} & \multicolumn{2}{c}{$9$} & \multicolumn{2}{c}{$15$} & \multicolumn{2}{c}{$20$} & \multicolumn{2}{c}{$30$} \\
\cmidrule(lr){2-3} \cmidrule(lr){4-5} \cmidrule(lr){6-7} \cmidrule(lr){8-9} \cmidrule(lr){10-11} \cmidrule(lr){12-13} \cmidrule(lr){14-15} \cmidrule(lr){16-17}
& D & C & D & C & D & C & D & C & D & C & D & C & D & C & D & C \\
\midrule
\multicolumn{17}{l}{\textit{Weight-based Methods}} \\
Task Arithmetic \cite{ilharco2023editing}                    & .547 & .715 & .493 & .723 & .458 & .675 & .468 & .698 & .568 & .747 & .522 & .724 & .515 & .727 & .502 & .720 \\
TIES \cite{yadav2023ties}                               & .566 & .745 & .484 & .720 & .451 & .679 & .467 & .711 & .572 & .749 & .523 & .730 & .517 & .727 & .494 & .713 \\
DARE \cite{yu2024dare}                               & .544 & .714 & .493 & .724 & .457 & .675 & .467 & .698 & .572 & .747 & .522 & .722 & .515 & .726 & .502 & .720 \\
RobustMerge \cite{zeng2025robustmerge}                       & .523 & .743 & .501 & .732 & .495 & .734 & .489 & .732 & .506 & .729 & .486 & .722 & .485 & .720 & .472 & .717 \\
KnOTS \cite{stoica2025model}                             & .579 & .760 & .510 & .737 & .504 & .731 & .487 & .725 & .541 & .740 & .523 & .733 & .506 & .730 & .500 & .728 \\
Core Space \cite{panariello2025accurate}                 & .603 & .757 & .516 & .732 & .505 & .725 & .492 & .703 & .599 & .759 & .551 & .740 & .542 & .739 & .529 & .740 \\
Iso-C \cite{marczak2025isoc}                              & .592 & .759 & .536 & .732 & .511 & .721 & .502 & .705 & .598 & .764 & .559 & .752 & .549 & .746 & .533 & .737 \\
\midrule
\multicolumn{17}{l}{\textit{Activation-based Methods}} \\
RegMean \cite{jin2022dataless}                      & .693 & \underline{.821} & \underline{.681} & .812 & \underline{.663} & .806 & .636 & .796 & .621 & .779 & .587 & .764 & .571 & .758 & .554 & .751 \\
Mix-of-Show \cite{gu2023mix}                 & .684 & .817 & .657 & .799 & .645 & .797 & .621 & .789 & .629 & .782 & .595 & .767 & .584 & .763 & .566 & .755 \\
SSR \cite{wei2026ssrmerge}                         & .649 & .798 & .628 & .792 & .619 & .791 & .607 & .788 & .589 & .768 & .574 & .759 & .571 & .760 & .569 & .757 \\
\rowcolor{gray!10}
SSR\,+\,$1/\gamma$ (ours)    & \underline{.697} & \underline{.821} & \textbf{.691} & \textbf{.820} & \textbf{.673} & \textbf{.816} & \underline{.674} & \underline{.815} & \underline{.646} & \underline{.788} & \underline{.640} & \underline{.781} & \underline{.632} & \underline{.778} & \underline{.626} & \textbf{.774} \\
\rowcolor{gray!10}
SSR\,+\,line search\ (ours)   & \textbf{.711} & \textbf{.828} & \underline{.681} & \underline{.819} & .657 & \underline{.808} & \textbf{.687} & \textbf{.819} & \textbf{.668} & \textbf{.789} & \textbf{.649} & \textbf{.783} & \textbf{.640} & \textbf{.780} & \textbf{.635} & \underline{.773} \\
\midrule
Merge-Aware (IterIS) \cite{chen2025iteris}      & .704 & .824 & .695 & .821 & .683 & .816 & .661 & .804 & .641 & .786 & .619 & .775 & .608 & .773 & .593 & .767 \\
\midrule
\UB{\textit{Individual LoRAs (avg)}}          & \UB{.723} & \UB{.833} & \UB{.723} & \UB{.833} & \UB{.723} & \UB{.833} & \UB{.723} & \UB{.833} & \UB{.671} & \UB{.804} & \UB{.671} & \UB{.804} & \UB{.671} & \UB{.804} & \UB{.671} & \UB{.804} \\
\bottomrule
\end{tabular}
%}
\caption{Subject fidelity for merging $K$ DreamBooth LoRAs on FLUX.1-dev and SD3.5-Medium, measured as DINOv2-B (D) and CLIP ViT-L/14 image (C) similarity to the reference images. Our methods are shaded gray; \textbf{bold}/\underline{underline} mark the best and second-best result per column (excluding IterIS). The upper bound (Individual LoRAs) is the average of the individually fine-tuned models.}
\label{tab:dreambooth}
\vspace{-10pt}
\end{table*}

{\noindent\textbf{Task.}}
We follow \cite{wei2026ssrmerge} and evaluate DreamBooth~\cite{ruiz2023dreambooth} as the main task. DreamBooth personalizes a text-to-image model like Flux1-dev or Stable Diffusion 3.5-Medium to a specific subject from a few reference images. The model is fine-tuned to bind a rare identifier token to the subject, so that a prompt such as \enquote{a sks dog on the beach} renders that particular dog in a new context. We train one LoRA~\cite{hu2022lora} per subject and merge $K$ of them into a single adapter. The merged model should preserve every subject it contains. For each subject, we therefore test whether it can still be generated faithfully when its identifier is used.

\mypara{Subjects}
In contrast to prior work, which evaluates subsets, we use all 30 subjects of the DreamBooth dataset~\cite{ruiz2023dreambooth}, i.e., we merge up to $K=30$ adapters. These include 21 objects (e.g., backpacks, plushies, sneakers, toys) and 9 live subjects (2 cats and 7 dogs). Each subject has 4 to 6 reference images. For each pool size $K \in \{3, 5, 9, 15, 20, 30\}$ and each of the 30 subjects, we merge the adapter of the target subject and $K{-}1$ other subjects drawn uniformly at random. The merge is evaluated on the target only, which gives 30 configurations per $K$. At $K{=}30$, every configuration contains the full pool.

\mypara{Adapters}
We train adapters for FLUX.1-dev and Stable Diffusion~3.5 Medium (SD3.5-M). We follow the DreamBooth LoRA recipe of \texttt{diffusers}: rank $r{=}32$ with $\alpha{=}r$, AdamW with a constant learning rate of $10^{-4}$ and weight decay $10^{-4}$, batch size 1, and train for 500 steps on $512{\times}512$ center crops, keeping text encoders frozen.

\mypara{Evaluation}
Images are generated at $512{\times}512$ resolution with 28 sampling steps, using a guidance scale of 3.5 for FLUX.1-dev and 5.0 for SD3.5-M. The test set has four prompts (\enquote{a photo of sks \aq{phrase}} and the subject on a wooden floor, on the beach, and in the snow). This gives 8 images per configuration. We measure subject fidelity with DINO and CLIP-I. We use DINOv2-B and CLIP ViT-L/14, respectively.

\mypara{Baselines}
We compare against three families of baselines where we tune all hyperparameters extensively (\cref*{sec:supp:hyperparameters}): (1) \emph{task-vector methods}, including Task Arithmetic \cite{ilharco2023editing} (scaled vector summation), TIES \cite{yadav2023ties} (magnitude-based pruning and sign consensus), and DARE \cite{yu2024dare} (random sparsification with rescaling); (2) \emph{low-rank methods}, including RobustMerge \cite{zeng2025robustmerge} (stabilizing singular directions via parameter pruning and rescaling), KnOTS \cite{stoica2025model} (joint SVD alignment), Core Space \cite{panariello2025accurate} (projection into shared subspace bases), and ISO-C \cite{marczak2025isoc} (uniform singular spectrum equalization); and (3) \emph{activation-based methods} that match task outputs on unlabeled data, including closed-form regression via RegMean \cite{jin2022dataless}, L-BFGS optimization via Mix-of-Show \cite{gu2023mix}, and iterative alignment on merged-model activations via IterIS \cite{chen2025iteris}. Activation-based methods are data-matched, see \cref*{sec:supp:calibrationdata}.

\subsection{Comparison to SOTA Methods on Dreambooth}

In \Cref{tab:dreambooth}, we show subject fidelity for merging 9 to 30 DreamBooth LoRAs on FLUX.1-dev and SD3.5-M, measured by DINOv2 (D) and CLIP image (C) similarity to the reference images. First, we observe that weight-space methods fall behind activation-based methods at every scale, and the gap widens as more adapters are merged. On FLUX, the best weight-space result at 9 adapters (Core Space, .603 D) is already behind RegMean by .090, and at 30 adapters the best weight-space score (Iso-C, .502) is .185 below SSR with line search. Task Arithmetic loses .079 D between 9 and 30 adapters, whereas our method loses .024. Second, rescaling turns SSR from a mid-ranking baseline into the strongest method. Plain SSR is behind RegMean and Mix-of-Show at small $K$, but adding $1/\gamma$ or line-search rescaling raises D at 30 adapters from .607 to .687 on FLUX and from .569 to .635 on SD3.5, with corresponding gains in C (.788 to .819 and .757 to .773). On SD3.5, SSR + Line Search obtains the best D score at every $K$. On FLUX, it is best at 9 and 30 adapters and within .010 of the best method elsewhere. Third, RegMean is competitive at small $K$ (.693 vs .697 D at 9 adapters on FLUX) and IterIS at 15 and 20 adapters (.695 vs .691 and .683 vs .673). IterIS follows a stronger paradigm, since it accounts for the other adapters during merging, but in a data-matched setting it is much more expensive (94min / 426 PFLOPS for IterIS vs.\ 7min / 21 PFLOPS for SSR + $1/\gamma$ when merging 30 adapters on SD3.5-M, see \cref*{sec:supp:runtime}), and at $K=30$ it falls behind our method by .026 D and .015 C. 
%Overall, we conclude that rescaling is greatly beneficial and achieves the best performance among compared methods.

\subsection{Results on Style Transfer and Concept Erasure}

{\noindent\textbf{WikiArt Style Transfer.}}
We build an art style transfer task using 20 artists from the WikiArt dataset \cite{tan2019improved}. For each artist, we sample 25 training and 20 reference images (see \cref*{sec:supp:wikiart}). LoRA adapters are trained using the same recipe as for DreamBooth, using \enquote{an image in the style of sks $\langle$code$\rangle$} as training prompt and \enquote{$\langle$content$\rangle$, in the style of sks $\langle$code$\rangle$} with four unseen content specifications as test prompts. Style fidelity is measured with CSD-spec \cite{somepalli2024measuring}, i.e., the mean cosine similarity to held-out images of the target style minus the mean similarity to the other 19 styles, and with the mean CLIP similarity to reference images.

In \cref{tab:style_and_erasure}, we show that unscaled SSR doesn't preserve distinct artistic styles well and achieves only low style specificity. Rescaling the merged adapter resolves this and brings substantial improvements. On FLUX.1-dev, SSR\,+\,line search raises CSD-spec from .018 to .078 at $K{=}9$ and from .015 to .085 at $K{=}20$. Scaling with $1/\gamma$ also yields gains, more than tripling the specificity of plain SSR at $K{=}20$. IterIS is the only competitive baseline, clearly outperforming RegMean, Mix-of-Show, and unscaled SSR. Nonetheless, it remains behind our methods, which achieve the best overall fidelity (.056 at $K{=}9$ and .027 at $K{=}20$).

\begin{table*}[t]
\centering
%\footnotesize
\setlength{\tabcolsep}{3pt}
%\resizebox{\textwidth}{!}{%
\begin{tabular}{l cccccccc ccccccccc}
\toprule
\multirow{4.5}{*}{\textbf{Method}} & \multicolumn{8}{c}{\textbf{WikiArt Style Transfer}} & \multicolumn{9}{c}{\textbf{CIFAR-10 Concept Erasure}} \\
\cmidrule(lr){2-9} \cmidrule(lr){10-18}
& \multicolumn{4}{c}{\textbf{FLUX.1-dev}} & \multicolumn{4}{c}{\textbf{SD3.5-Medium}} & \multicolumn{9}{c}{\textbf{SD~v1.4}} \\
\cmidrule(lr){2-5} \cmidrule(lr){6-9} \cmidrule(lr){10-18}
& \multicolumn{2}{c}{$9$} & \multicolumn{2}{c}{$20$} & \multicolumn{2}{c}{$9$} & \multicolumn{2}{c}{$20$} & \multicolumn{3}{c}{$3$} & \multicolumn{3}{c}{$5$} & \multicolumn{3}{c}{$7$} \\
\cmidrule(lr){2-3} \cmidrule(lr){4-5} \cmidrule(lr){6-7} \cmidrule(lr){8-9} \cmidrule(lr){10-12} \cmidrule(lr){13-15} \cmidrule(lr){16-18}
& S & C & S & C & S & C & S & C & E & P & $H_o$ & E & P & $H_o$ & E & P & $H_o$ \\
\midrule
RegMean \cite{jin2022dataless}             & .052 & \underline{.515} & .041 & .513 & .015 & .532 & .005 & .538 & .83 & .98 & .828 & .84 & .99 & .839 & .82 & .99 & .819 \\
Mix-of-Show \cite{gu2023mix}              & .025 & .490 & .014 & .488 & .022 & .534 & .008 & .535 & .86 & .98 & \underline{.869} & .86 & .94 & .866 & .86 & .88 & .843 \\
SSR \cite{wei2026ssrmerge}                & .018 & .489 & .015 & .489 & .007 & .531 & .003 & .538 & .82 & .99 & .826 & .86 & .93 & .862 & .85 & .91 & .846 \\
\rowcolor{gray!10}
SSR\,+\,$1/\gamma$ (ours)                 & {.053} & .506 & \underline{.055} & \underline{.514} & .029 & \underline{.537} & \textbf{.027} & \underline{.545} & .87 & .98 & \underline{.869} & .88 & .92 & \underline{.879} & .88 & .86 & \textbf{.852} \\
\rowcolor{gray!10}
SSR\,+\,line search (ours)                & \textbf{.078} & \textbf{.556} & \textbf{.085} & \textbf{.564} & \textbf{.056} & \textbf{.576} & .019 & \textbf{.547} & .91 & .97 & \textbf{.923} & .89 & .90 & \textbf{.885} & .89 & .81 & \underline{.848} \\
\midrule
Merge-Aware (IterIS) \cite{chen2025iteris}& \underline{.056} & .490 & .045 & .498 & \underline{.036} & .535 & \underline{.020} & .532 & .83 & .96 & .861 & .86 & .88 & .859 & .86 & .77 & .820 \\
\midrule
\UB{\textit{Individual LoRAs / erasers}}  & \UB{.120} & \UB{.523} & \UB{.120} & \UB{.523} & \UB{.140} & \UB{.558} & \UB{.140} & \UB{.558} & \UB{.86} & \UB{.99} & \UB{.919} & \UB{.86} & \UB{.99} & \UB{.918} & \UB{.86} & \UB{.98} & \UB{.916} \\
\bottomrule
\end{tabular}
%}
\caption{Style fidelity and concept erasure results across varying adapter counts $K$. \emph{Left (WikiArt)}: Merging $K \in \{9, 20\}$ style LoRAs on FLUX.1-dev and SD3.5-Medium, evaluated by CSD-spec (S) and CLIP image similarity (C) to held-out references. \emph{Right (Concept Erasure)}: Merging $K \in \{3, 5, 7\}$ MACE object erasers for CIFAR-10 classes on SD~v1.4, evaluated by erasure efficacy E ($1{-}\mathrm{Acc}_e$), preservation P ($\mathrm{Acc}_s$), and their harmonic mean $H_o$. \textbf{Bold}/\underline{underline} highlight the best and second-best results per column.}
\label{tab:style_and_erasure}
\end{table*}

{\noindent\textbf{Concept Erasure.}}
We merge adapters that remove concepts, following the CIFAR-10 protocol of MACE~\cite{lu2024mace} on SD~v1.4. We train one MACE eraser per class, convert it into an exact LoRA of the cross-attention projections, and merge $K \in \{3, 5, 7\}$ erasers of random classes. A CLIP classifier measures erasure efficacy E on the erased classes, preservation P on the remaining classes, and $H_o$, the harmonic mean of E, P, and generality to synonyms. Calibration uses only text prompts. Following MACE, the router includes a preservation term on 30 neutral concepts.

In \cref{tab:style_and_erasure}, SSR and the baselines erase almost as well as the individual erasers (E .82 to .86) and keep preservation high. Rescaling raises erasure at every $K$. With $1/\gamma$ ($\gamma$ between .91 and .92), SSR improves from .826 to .869 $H_o$ at $K{=}3$, from .862 to .879 at $K{=}5$, and from .846 to .852 at $K{=}7$, where it obtains the best $H_o$. SSR\,+\,line search obtains the best $H_o$ at $K{=}3$ and $K{=}5$ (.923 and .885, compared with .869 and .866 for Mix-of-Show) and even erases more than the individual erasers at $K{=}3$ (E .91 vs.\ .86). It selects a larger $\alpha$ than $1/\gamma$ (1.17 to 1.52), i.e., $s^\star > 1$ in \cref{eq:theory:downstream:alpha_star}: the metric rewards erasing more strongly than each individual eraser. In \cref{fig:qualitative}\,(c), RegMean and plain SSR still generate the erased bird, cat, and frog, while SSR\,+\,line search removes all three and keeps the dog; Task Arithmetic erases but also destroys the kept class. Weight-based baselines are reported in \cref*{sec:supp:erasure}.

\subsection{Qualitative Examples for Image Generation}
\label{sec:qualitative}

\begin{figure*}[t]
\centering
\includegraphics[width=\textwidth]{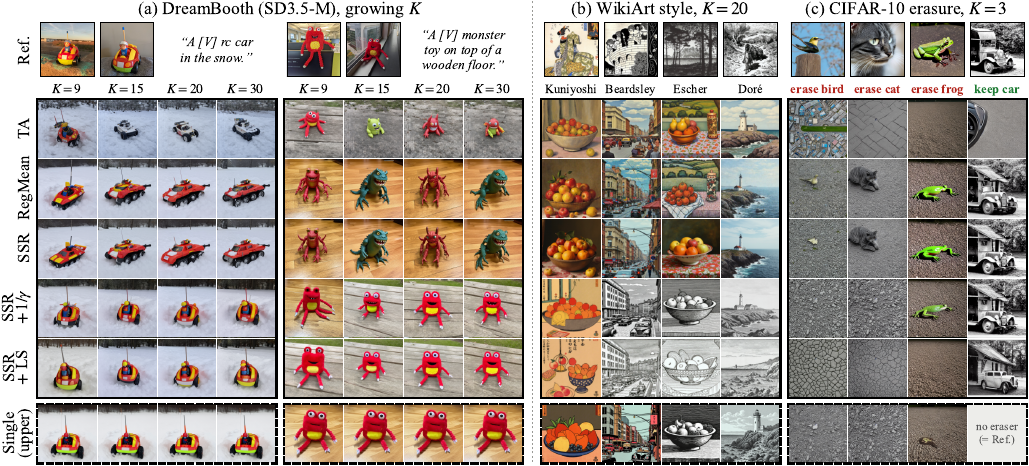}
\caption{Qualitative results for (a) DreamBooth subjects on SD3.5-M for growing $K$, (b) WikiArt styles at $K{=}20$ on FLUX.1-dev, and (c) CIFAR-10 concept erasure at $K{=}3$ on SD~v1.4. The bottom row shows each adapter alone (Single) as an upper bound. TA, RegMean, and plain SSR lose the subjects and drift towards generic styles; in erasure, RegMean and SSR still generate the erased classes, and TA also destroys the kept class. Rescaling SSR by $1/\gamma$ or line search (LS) recovers subjects and styles, and LS removes all three erased classes.}
\label{fig:qualitative}
\end{figure*}

\cref{fig:qualitative} illustrates the quantitative trends. On DreamBooth (a), the gap between methods widens as the number of merged adapters grows. Task Arithmetic (TA) loses the subject early and at $K=30$ produces a generic toy car or monster with no trace of the reference. RegMean and plain SSR retain some attributes, such as the red color of the car, but the subject gradually changes shape and identity as $K$ increases. Rescaling SSR by $1/\gamma$ or by $\alpha$ found by line search keeps both subjects close to the single-adapter reference across all $K$, including the full pool of 30 adapters. This matches our analysis: the merged output is a shrunken copy of the target, and undoing the shrinkage recovers the subject even when many adapters overlap.
On WikiArt (b), TA, RegMean, and plain SSR drift towards a shared, generic painterly style, whereas the rescaled variants reproduce the distinct styles of Kuniyoshi, Beardsley, Escher, and Dor\'e. On CIFAR-10 (c), RegMean and SSR still generate the erased classes, and TA also damages the kept class. Scaled SSR removes all erased classes while preserving the car.

\subsection{Performance on LLM and Classification Tasks}
\label{sec:ablations:othertasks}

{\noindent \textbf{Setting.}}
We test whether rescaling merged adapters also improves performance on tasks other than image generation. Specifically, we test image classification, text classification, and text generation. \textit{Image Classification} includes 20 image-classification tasks from \cite{stoica2025model}. Here, we train a LoRA adapter for each task on the CLIP ViT-B/32 and ViT-L/14 \cite{radford2021learning} image encoders and measure performance by accuracy.
\textit{LLM-NLI} and \textit{LLM-gen} include 6 natural language inference tasks from \cite{stoica2025model} (LLM-NLI) and 3 generative tasks (math$\to$GSM8K \cite{cobbe2021training}, code$\to$MBPP \cite{austin2021program}, instruction following$\to$IFEval \cite{zhou2023instruction}). We train LoRAs for Llama-3.1-8B \cite{grattafiori2024llama} on the provided NLI data for LLM-NLI and on the following datasets for LLM-gen: math$\to$MetaMath \cite{yu2024metamath}, code$\to$Magicoder \cite{wei2024magicoder}, and instruction following$\to$Tülu 3 SFT IF data \cite{lambert2025tulu}. LLM-NLI is evaluated by accuracy, and LLM-gen by the respective task metrics (exact match, pass@1, and the official scorer from lm-eval-harness for IFEval).
\textit{LoRAHub} includes 20 adapters selected from \cite{huang2024lorahub}. The adapters are trained for FLAN-T5 \cite{chung2024scaling} one of 11 English NLP datasets, which cover question answering and binary judgement tasks like sentiment. Performance is measured by exact match accuracy.

\begin{figure*}
\centering
\includegraphics[width=\linewidth]{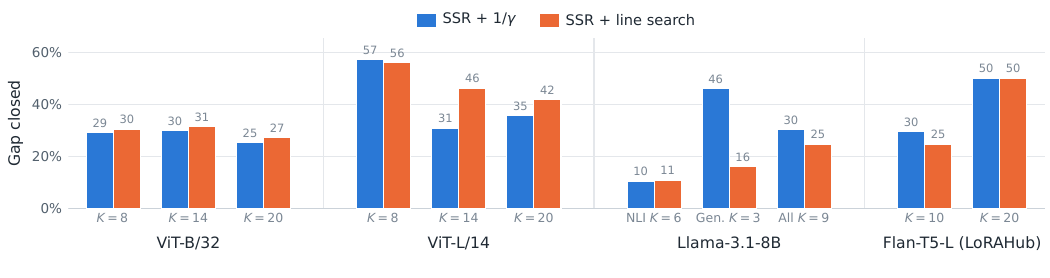}
\caption{Percentage of the gap to the fine-tuned upper bound closed by SSR with $1/\gamma$ and line-search rescaling for image classification (CLIP ViT-B/32, ViT-L/14), language understanding and generation (Llama-3.1-8B), and LoRAHub. $K$ is the number of merged adapters.}
\label{fig:ablations:othertasks}
\end{figure*}

\mypara{Results}
We show results in \cref{fig:ablations:othertasks}. For each task and configuration, we report the percentage of the gap from plain SSR ($\alpha = 1$) to the fine-tuned upper bound that is closed by the merged model. Here, a score of 50 means that half of the gap is closed. Across domains, rescaling consistently yields substantial performance gains. On image classification with ViT-B/32, rescaling closes 25\% to 31\% of the gap across all adapter counts and improves accuracy from 68.4\% to 73.8\% at $K{=}20$. On ViT-L/14, the improvements are even more pronounced, recovering up to 57\% of the gap at $K{=}8$ (improving accuracy from 83.9\% to 89.1\%) and 36\%--42\% at $K{=}20$ (from 77.3\% to 83.7\%). On LoRAHub, rescaling recovers 25\%--30\% of the gap at $K{=}10$ and 50\% at $K{=}20$. Even across all 9 Llama-3.1 tasks combined, rescaling closes 25\% to 30\% of the gap.

Scores on LLMs are lower, especially for NLI. We also oberve that performance on generative tasks is mainly driven by IFEval, and Line Search underperforms here. We analyzed the reasons and found that (i) optimal rescaling factors for the three generative tasks are anticorrelated, so optimal performance for one task yields worst performance for the others. IFEval yields the largest gains overall, so it dominates the merge. (ii) Performance on the validation set is uncorrelated with performance on the test set, which explains why Line Search fails to select a good $\alpha$. Since these problems are specific to LLM generative tasks and this is not the focus of our work, we leave further exploration to future research. Finally, we also want to clarify that rescaled SSR does not achieve state-of-the-art performance on all of these tasks, but our claim is that rescaling substantially improves SSR and other OLS-based methods.

\begin{table*}[t]
\centering
%\footnotesize
\setlength{\tabcolsep}{3pt}
\resizebox{\textwidth}{!}{%
\begin{tabular}{l cc cccccc ccc ccc ccc}
\toprule
\multirow{4.5}{*}{\textbf{Method}} & \multicolumn{8}{c}{\textbf{DreamBooth}} & \multicolumn{6}{c}{\textbf{Image Classification}} & \multicolumn{3}{c}{\textbf{LLM}} \\
\cmidrule(lr){2-9} \cmidrule(lr){10-15} \cmidrule(lr){16-18}
& \multicolumn{2}{c}{FLUX.1-dev} & \multicolumn{6}{c}{SD3.5-Medium} & \multicolumn{3}{c}{ViT-B/32} & \multicolumn{3}{c}{ViT-L/14} & \multicolumn{3}{c}{Llama-3.1-8B} \\
\cmidrule(lr){2-3} \cmidrule(lr){4-9} \cmidrule(lr){10-12} \cmidrule(lr){13-15} \cmidrule(lr){16-18}
& \multicolumn{2}{c}{$30$} & \multicolumn{2}{c}{$9$} & \multicolumn{2}{c}{$15$} & \multicolumn{2}{c}{$30$} & $8$ & $14$ & $20$ & $8$ & $14$ & $20$ & NLI ($6$) & Gen ($3$) & Comb ($9$) \\
\cmidrule(lr){2-3} \cmidrule(lr){4-5} \cmidrule(lr){6-7} \cmidrule(lr){8-9} \cmidrule(lr){10-15} \cmidrule(lr){16-18}
& D & C & D & C & D & C & D & C & \multicolumn{6}{c}{Accuracy (\%)} & \multicolumn{3}{c}{Score (\%)} \\
\midrule
RegMean                  & .646 & .800 & .619 & .778 & .587 & .762 & .553 & .751 & 82.5 & 74.8 & 71.7 & 89.5 & 84.4 & 81.0 & \textbf{86.3} & 61.0 & 77.0 \\
\rowcolor{gray!05}
\quad $+\,1/\gamma$            & \textbf{.693} & \textbf{.822} & {.644} & {.789} & {.635} & {.779} & {.600} & {.766} & \textbf{84.1} & {79.0} & \textbf{76.2} & {91.4} & {88.6} & {86.7} & 86.0 & {61.8} & {77.8} \\
\rowcolor{gray!05}
\quad $+\,$line search         & .653 & .815 & \textbf{.663} & \textbf{.790} & \textbf{.655} & \textbf{.785} & \textbf{.623} & \textbf{.767} & 84.0 & \textbf{79.1} & 76.1 & \textbf{91.4} & \textbf{88.7} & \textbf{86.8} & 86.1 & \textbf{64.2} & \textbf{77.9} \\
\midrule
Mix-of-Show              & .621 & .792 & .629 & .783 & .592 & .768 & .566 & .755 & 83.1 & 75.7 & 72.4 & 89.8 & 84.9 & 81.7 & \textbf{86.2} & 61.4 & 77.1 \\
\rowcolor{gray!05}
\quad $+\,1/\gamma$            & \textbf{.669} & \textbf{.812} & {.650} & \textbf{.790} & {.639} & \textbf{.780} & {.606} & {.767} & \textbf{84.3} & \textbf{79.4} & \textbf{76.5} & \textbf{91.5} & {88.7} & \textbf{87.0} & 86.0 & {62.5} & \textbf{78.5} \\
\rowcolor{gray!05}
\quad $+\,$line search         & .635 & .806 & \textbf{.661} & .789 & \textbf{.645} & \textbf{.780} & \textbf{.622} & \textbf{.770} & 84.2 & \textbf{79.4} & 76.4 & 91.4 & \textbf{88.8} & \textbf{87.0} & 86.1 & \textbf{63.6} & 77.8 \\
\midrule
SSR                      & .607 & .788 & .589 & .768 & .574 & .759 & .569 & .757 & 77.3 & 70.4 & 68.4 & 84.7 & 79.9 & 77.3 & 85.1 & 58.2 & 74.2 \\
\rowcolor{gray!05}
\quad $+\,1/\gamma$            & .674 & .815 & .646 & .788 & .640 & .781 & .626 & .774 & 81.0 & 75.7 & 73.4 & \textbf{89.7} & 85.8 & 82.8 & \textbf{86.0} & \textbf{61.5} & \textbf{77.1} \\
\rowcolor{gray!05}
\quad $+\,$line search         & \textbf{.687} & \textbf{.819} & \textbf{.668} & \textbf{.789} & \textbf{.652} & \textbf{.786} & \textbf{.636} & \textbf{.777} & \textbf{81.1} & \textbf{75.9} & \textbf{73.8} & 89.6 & \textbf{86.2} & \textbf{83.7} & \textbf{86.0} & 59.4 & 76.6 \\
\midrule
\UB{\textit{Individual LoRAs}}      & \UB{.723} & \UB{.833} & \UB{.671} & \UB{.804} & \UB{.671} & \UB{.804} & \UB{.671} & \UB{.804} & \UB{88.2} & \UB{88.2} & \UB{88.2} & \UB{93.2} & \UB{92.6} & \UB{92.7} & \UB{93.2} & \UB{65.4} & \UB{83.9} \\
\bottomrule
\end{tabular}
}
\caption{Evaluation of rescaling on OLS-based merging methods (SSR, RegMean and Mix-of-Show) for DreamBooth, image classification, and Llama-3.1-8B. \textbf{Bold} highlight best result within each method. \textit{Individual LoRAs} means the fine-tuned upper bound.}
\label{tab:othermethods}
\end{table*}

\subsection{Scaling Gains on RegMean and Mix-of-Show}
\label{sec:ablations:othermethods}

{\noindent \textbf{Setting.}}
We test whether adapter rescaling is helpful on RegMean \cite{jin2022dataless} and Mix-of-Show \cite{gu2023mix}. Both methods solve the following general problem:
\begin{equation}
    W^{\star} = \arg\min_W \sum_k\|(W_0 + \Delta W_k)X_k - WX_k\|_F^2,
\end{equation}
where $X_i$ is the activation of task $k$ and $\Delta W_k$ the corresponding adapter. As these methods also optimize an OLS objective, the resulting merged adapters $W^{\star}$ suffer from shrinkage, too.

\mypara{Results}
Performance gains of rescaling for RegMean and Mix-of-Show are in \cref{tab:othermethods}. We include the DreamBooth task, Image Classification, and Llama-3.1 finetuning. Rescaling improves performance across almost all settings, for example raising image classification accuracy on ViT-L/14 from 81.0\% to 86.7\% for RegMean and DreamBooth DINO on SD3.5-M at $K{=}30$ from .553 to .600, with only slight drops on Llama NLI (e.g., 86.3\% to 86.0\% for RegMean). Comparing the two base methods, Mix-of-Show obtains slightly higher unscaled accuracy than RegMean on classification (e.g., 72.4\% vs.\ 71.7\% on ViT-B/32 at $K{=}20$), but both methods show consistent and comparable improvements from rescaling. Comparing the two rescaling approaches, line search achieves better results on SD3.5-M (e.g., .623 vs .600 DINO for RegMean at $K{=}30$) and on Llama generation (64.2\% vs.\ 61.8\%), whereas $1/\gamma$ performs better on FLUX.1-dev (.693 vs .653 DINO) and closely matches line search on image classification (76.2\% vs.\ 76.1\% on ViT-B/32 at $K{=}20$). This agrees with our findings in \cref{sec:ablations:othertasks} where gains on LLMs are weaker. We think that LLM merging is substantially different and deserves independent investigation.

\section{Conclusion}
This paper demonstrates and analyzes shrinkage as a problem in OLS-based model merging methods. We show that methods like SSR \cite{wei2026ssrmerge}, RegMean \cite{jin2022dataless}, and Mix-of-Show \cite{gu2023mix} can achieve substantially improved performance on image generation tasks like few-shot subject-driven generation, style transfer, and concept erasue, by rescaling merged adapters. We also theoretically derive a strong heuristic for this factor. Beyond image generation, rescaling enables gains on image classification, text classification, and LLM tasks. Together, our experiments suggest that the problem as well as the proposed fix are universal to OLS-based merging. Beyond the specific performance improvements, our findings motivate re-evaluating the trade-off between local merging targets and downstream effects, which may not always be aligned. Our analysis of OLS shrinkage is a first step toward efficient merging while taking global effects into account.

\mypara{Limitations}
While rescaling adapters fitted with OLS yields substantial improvements especially on image generation tasks, it remains a post-hoc method that mitigates shrinkage but doe not solve it during merging. Future research should develop methods to directly achieve this in one step. Our $1/\gamma$ heuristic for the scaling factor works well, but calibration sometimes fails when $s^\star$ is far from $1$. Finally, like almost all merging methods, we don't consider activation drift in the merged model across layers, which is an important next step. Finally, we find results on LLM merging weaker empirically, so better understanding this setting is also interesting but out of scope of our work.

\section*{LLM Usage Statement}
All contributions presented in this paper are our own. For running experiments, we used LLMs to support our research. Specifically, we used an LLM (Claude) to implement methods and coordinate experiments. Furthermore, the formalization of our findings into a coherent derivation (\cref{sec:theory}, \cref*{sec:supp:detailederivation}) was supported by different LLMs (Qwen, Gemini). We used Qwen, Claude, and Gemini to draft individual sections of the paper which were then revised and integrated by the authors. We take full responsibility for all contents of the paper.

% Bibliography
{
    \small
    \bibliographystyle{ieeenat_fullname}
    \bibliography{main}
}

% --- Right after your references ---
\clearpage
\onecolumn % Switch to single column

% Redefine \maketitlesupplementary by stripping out \twocolumn[...]
\renewcommand{\maketitlesupplementary}{%
  \newpage
  {%
    \centering
    \Large
    \textbf{\thetitle}\\
    \vspace{0.5em}Supplementary Material \\
    \vspace{1.0em}
    \par
  }%
}

% Print supplementary title
\maketitlesupplementary

% Start appendix
\appendix
\crefalias{section}{appendix}
\crefalias{subsection}{appendix}
\crefalias{subsubsection}{appendix}

\section{Detailed Derivations of Theoretical Claims}
\label{sec:supp:detailederivation}

Below we provide step-by-step derivations of key identities in \cref*{sec:theory}. We restate all definitions first.

\vspace{1em}
\hrule
\vspace{1em}

\subsection*{Notation and Definitions}

For $k = 1, \dots, K$, let $Y_k \in \mathbb{R}^{d \times m}$ be the target output of the adapter trained for task-$k$ in isolation, and let $\hat{Y}_k$ be the prediction of the merged router. We decompose
\begin{equation}
\hat{Y}_k = \gamma\, Y_k + U_k,
\label{eq:supp:decomposition}
\end{equation}
where the scalar $\gamma$ and the residual $U_k$ are defined so that the orthogonality condition
\begin{equation}
\sum_{k=1}^{K} \langle U_k,\, Y_k \rangle_F = 0
\label{eq:supp:orthogonality}
\end{equation}
holds. Concretely, $\gamma$ is chosen as the orthogonal projection coefficient:
\begin{equation}
\gamma = \frac{\displaystyle\sum_{k=1}^{K} \langle \hat{Y}_k,\, Y_k \rangle_F}{\displaystyle\sum_{k=1}^{K} \|Y_k\|_F^2}.
\label{eq:supp:gamma}
\end{equation}
We further define the normalised quantities
\begin{equation}
\hat{h} = \frac{\displaystyle\sum_{k=1}^{K} \|\hat{Y}_k\|_F^2}{\displaystyle\sum_{k=1}^{K} \|Y_k\|_F^2}, \qquad e = \frac{\displaystyle\sum_{k=1}^{K} \|U_k\|_F^2}{\displaystyle\sum_{k=1}^{K} \|Y_k\|_F^2}.
\label{eq:supp:defs}
\end{equation}
Throughout, $\langle A, B\rangle_F = \tr(A^\top B)$ denotes the Frobenius inner product and $\|A\|_F^2 = \langle A, A\rangle_F$.
For readability, write
\begin{equation}
S \coloneqq \sum_{k=1}^{K} \|Y_k\|_F^2 > 0
\label{eq:supp:S-def}
\end{equation}
for the total target energy. All normalised quantities are ratios with $S$ in the denominator.

\vspace{1em}
\hrule
\vspace{1em}

\subsection*{Derivation 1: $\;\hat{h} = \gamma^2 + e$}

\paragraph{Goal.} Starting from the decomposition $\hat{Y}_k = \gamma Y_k + U_k$ and the orthogonality condition~\cref{eq:supp:orthogonality}, prove
\begin{equation}
\hat{h} = \gamma^2 + e.
\label{eq:supp:d1-goal}
\end{equation}

\paragraph{Step 1. Expand the squared norm of $\hat{Y}_k$.}
For each $k$,
\begin{equation}
\|\hat{Y}_k\|_F^2 = \|\gamma Y_k + U_k\|_F^2.
\label{eq:supp:d1-norm}
\end{equation}
By bilinearity of the Frobenius inner product (i.e., $\|A+B\|_F^2 = \|A\|_F^2 + 2\langle A,B\rangle_F + \|B\|_F^2$),
\begin{equation}
\|\hat{Y}_k\|_F^2 = \gamma^2\|Y_k\|_F^2 + 2\gamma\,\langle Y_k,\, U_k\rangle_F + \|U_k\|_F^2.
\label{eq:supp:d1-expand}
\end{equation}

\paragraph{Step 2. Sum over all tasks.}
Summing~\cref{eq:supp:d1-expand} from $k=1$ to $K$:
\begin{equation}
\sum_{k=1}^K \|\hat{Y}_k\|_F^2 = \gamma^2 \sum_{k=1}^K \|Y_k\|_F^2 + 2\gamma \sum_{k=1}^K \langle Y_k,\, U_k\rangle_F + \sum_{k=1}^K \|U_k\|_F^2.
\label{eq:supp:d1-sum}
\end{equation}

\paragraph{Step 3. Apply the orthogonality condition.}
By condition~\cref{eq:supp:orthogonality}, the middle sum vanishes:
\begin{equation}
\sum_{k=1}^K \langle Y_k,\, U_k\rangle_F = 0.
\label{eq:supp:d1-ortho-vanish}
\end{equation}
Therefore~\cref{eq:supp:d1-sum} simplifies to
\begin{equation}
\sum_{k=1}^K \|\hat{Y}_k\|_F^2 = \gamma^2 \sum_{k=1}^K \|Y_k\|_F^2 + \sum_{k=1}^K \|U_k\|_F^2.
\label{eq:supp:d1-simplified}
\end{equation}

\paragraph{Step 4. Divide by $S$ and recognise the definitions.}
Dividing both sides of~\cref{eq:supp:d1-simplified} by $S = \sum_k \|Y_k\|_F^2$:
\begin{equation}
\frac{\sum_k \|\hat{Y}_k\|_F^2}{S} = \gamma^2 \cdot \frac{\sum_k \|Y_k\|_F^2}{S} + \frac{\sum_k \|U_k\|_F^2}{S}.
\label{eq:supp:d1-divide}
\end{equation}
The left-hand side is $\hat{h}$ by definition~\cref{eq:supp:defs}. The first term on the right is $\gamma^2 \cdot 1 = \gamma^2$. The second term is $e$ by definition~\cref{eq:supp:defs}. Hence
\begin{equation}
\boxed{\hat{h} = \gamma^2 + e.}
\label{eq:supp:d1-result}
\end{equation}
\hfill $\blacksquare$

\vspace{1em}
\hrule
\vspace{1em}

\subsection*{Derivation 2: $\;\displaystyle\sum_{k=1}^{K} \|c\hat{Y}_k - Y_k\|_F^2 = \bigl(c^2\hat{h} - 2c\gamma + 1\bigr)\,S$}

\paragraph{Goal.} For any scalar $c \in \mathbb{R}$, expand the reconstruction objective $\phi(c) = \sum_k \|c\hat{Y}_k - Y_k\|_F^2$ and express it in terms of $\hat{h}$, $\gamma$, and $S$.

\paragraph{Step 1. Expand the squared norm for a single task.}
For each $k$,
\begin{equation}
\|c\hat{Y}_k - Y_k\|_F^2 = \langle c\hat{Y}_k - Y_k,\; c\hat{Y}_k - Y_k \rangle_F.
\label{eq:supp:d2-inner}
\end{equation}
Expanding by bilinearity:
\begin{equation}
= c^2\|\hat{Y}_k\|_F^2 - 2c\,\langle \hat{Y}_k,\, Y_k\rangle_F + \|Y_k\|_F^2.
\label{eq:supp:d2-expand}
\end{equation}

\paragraph{Step 2. Sum over all tasks.}
\begin{equation}
\sum_{k=1}^K \|c\hat{Y}_k - Y_k\|_F^2 = c^2 \sum_{k=1}^K \|\hat{Y}_k\|_F^2 - 2c \sum_{k=1}^K \langle \hat{Y}_k,\, Y_k\rangle_F + \sum_{k=1}^K \|Y_k\|_F^2.
\label{eq:supp:d2-sum}
\end{equation}

\paragraph{Step 3. Rewrite each sum using the normalised definitions.}
\begin{itemize}
    \item By definition of $\hat{h}$ in~\cref{eq:supp:defs}: $\;\displaystyle\sum_{k=1}^K \|\hat{Y}_k\|_F^2 = \hat{h}\, S.$
    \item By definition of $\gamma$ in~\cref{eq:supp:gamma}: $\;\displaystyle\sum_{k=1}^K \langle \hat{Y}_k,\, Y_k\rangle_F = \gamma\, S.$
    \item By definition of $S$: $\;\displaystyle\sum_{k=1}^K \|Y_k\|_F^2 = S.$
\end{itemize}

\paragraph{Step 4. Substitute into \cref{eq:supp:d2-sum}.}
\begin{equation}
\sum_{k=1}^K \|c\hat{Y}_k - Y_k\|_F^2 = c^2 \hat{h}\,S - 2c\,\gamma\,S + S = \bigl(c^2\hat{h} - 2c\gamma + 1\bigr)\,S.
\label{eq:supp:d2-substitute}
\end{equation}
Therefore
\begin{equation}
\boxed{\phi(c) = \sum_{k=1}^{K} \|c\hat{Y}_k - Y_k\|_F^2 = \bigl(c^2\hat{h} - 2c\gamma + 1\bigr)\sum_{k=1}^{K}\|Y_k\|_F^2.}
\label{eq:supp:d2-result}
\end{equation}
\hfill $\blacksquare$

\vspace{1em}
\hrule
\vspace{1em}

\subsection*{Derivation 3: $\;\phi'(1)=0 \;\Longrightarrow\; \displaystyle\sum_{k=1}^{K}\langle \hat{Y}_k - Y_k,\;\hat{Y}_k\rangle_F = 0$}

\paragraph{Goal.} Show that, since $R^\star$ is the global OLS minimiser, i.e., $c=1$ minimizes $\phi$, the stated inner-product identity holds.

\paragraph{Step 1.} Write
\begin{equation}
\phi(c) = \sum_{k=1}^K \|c\hat{Y}_k - Y_k\|_F^2 = \sum_{k=1}^K \langle c\hat{Y}_k - Y_k,\; c\hat{Y}_k - Y_k\rangle_F.
\label{eq:supp:d3-phi}
\end{equation}

\paragraph{Step 2.} Differentiate under the sum. For a single term, using the product rule for the inner product:
\begin{equation}
\frac{d}{dc}\|c\hat{Y}_k - Y_k\|_F^2 = 2\,\langle \hat{Y}_k,\; c\hat{Y}_k - Y_k\rangle_F.
\label{eq:supp:d3-deriv}
\end{equation}
(To see this: let $f(c) = c\hat{Y}_k - Y_k$. Then $\frac{d}{dc}\langle f, f\rangle_F = 2\langle f', f\rangle_F = 2\langle \hat{Y}_k, c\hat{Y}_k - Y_k\rangle_F$.)

\paragraph{Step 3.} Sum over $k$ and evaluate at $c = 1$:
\begin{equation}
\phi'(1) = 2\sum_{k=1}^K \langle \hat{Y}_k,\; \hat{Y}_k - Y_k\rangle_F = 2\sum_{k=1}^K \langle \hat{Y}_k - Y_k,\; \hat{Y}_k\rangle_F,
\label{eq:supp:d3-eval}
\end{equation}
where the last equality uses symmetry of the Frobenius inner product: $\langle A, B\rangle_F = \langle B, A\rangle_F$.

\paragraph{Step 4.} Setting $\phi'(1) = 0$ and dividing by 2:
\begin{equation}
\boxed{\sum_{k=1}^{K} \langle \hat{Y}_k - Y_k,\;\hat{Y}_k\rangle_F = 0.}
\label{eq:supp:d3-result}
\end{equation}
\hfill $\blacksquare$

\vspace{1em}
\hrule
\vspace{1em}

\subsection*{Derivation 4: $\;\displaystyle\sum_{k=1}^{K}\langle \hat{Y}_k - Y_k,\;\hat{Y}_k\rangle_F = 0 \;\Longrightarrow\; \hat{h} = \gamma$}

\paragraph{Goal.} Starting from the identity established in Derivation 3, derive $\hat{h} = \gamma$ using only the definitions of $\hat{h}$ and $\gamma$.

\paragraph{Step 1. Expand the inner product by linearity in the first argument.}
For each $k$:
\begin{equation}
\langle \hat{Y}_k - Y_k,\;\hat{Y}_k\rangle_F = \langle \hat{Y}_k,\;\hat{Y}_k\rangle_F - \langle Y_k,\;\hat{Y}_k\rangle_F = \|\hat{Y}_k\|_F^2 - \langle Y_k,\;\hat{Y}_k\rangle_F.
\label{eq:supp:d4-expand}
\end{equation}

\paragraph{Step 2. Sum over $k$.}
\begin{equation}
\sum_{k=1}^K \langle \hat{Y}_k - Y_k,\;\hat{Y}_k\rangle_F = \sum_{k=1}^K \|\hat{Y}_k\|_F^2 - \sum_{k=1}^K \langle Y_k,\;\hat{Y}_k\rangle_F.
\label{eq:supp:d4-sum}
\end{equation}

\paragraph{Step 3. Identify each sum with the normalized definitions.}
\begin{itemize}
    \item The first sum: $\;\displaystyle\sum_{k=1}^K \|\hat{Y}_k\|_F^2 = \hat{h}\,S$ by definition~\cref{eq:supp:defs}.
    \item The second sum: $\;\displaystyle\sum_{k=1}^K \langle Y_k,\;\hat{Y}_k\rangle_F = \sum_{k=1}^K\langle \hat{Y}_k,\;Y_k\rangle_F = \gamma\,S$ by definition~\cref{eq:supp:gamma} and symmetry of $\langle\cdot,\cdot\rangle_F$.
\end{itemize}

\paragraph{Step 4. Substitute into \cref{eq:supp:d4-sum} and set equal to zero.}
\begin{equation}
0 = \hat{h}\,S - \gamma\,S = (\hat{h} - \gamma)\,S.
\label{eq:supp:d4-sub}
\end{equation}

\paragraph{Step 5. Conclude.}
Since $S = \sum_k \|Y_k\|_F^2 > 0$ (the targets are not all zero), we can divide by $S$:
\begin{equation}
\boxed{\hat{h} = \gamma.}
\label{eq:supp:d4-result}
\end{equation}
\hfill $\blacksquare$

\vspace{1em}
\hrule
\vspace{1em}

\subsection*{Derivation 5: $\;\mathcal{L}(\alpha) \approx \mathcal{L}_{\min} + \frac{a}{2}(\alpha\gamma - s^\ast)^2 + \frac{b}{2}\alpha^2 e$}

\paragraph{Goal.} Starting from the decomposition $\hat{Y}_k = \gamma Y_k + U_k$ and the observation that the global task loss $\mathcal{L}$ is minimised at factor $s^\star$ (with $s^\star \approx 1$ for well-trained LoRAs), derive the second-order approximation of $\mathcal{L}$ as a function of the global rescaling factor $\alpha$ applied to the OLS router $R^\star$.

\paragraph{Step 1. Identify the how $\alpha$ affects the loss.}
Scaling the router by $\alpha$ maps the prediction for task $k$ to
\begin{equation}
\alpha\,\hat{Y}_k = \underbrace{\alpha\gamma}_{\coloneqq\; s(\alpha)}\, Y_k \;+\; \alpha\, U_k.
\label{eq:supp:d5-scaling}
\end{equation}
Thus $\alpha$ controls two quantities simultaneously:
\begin{itemize}
    \item the \emph{signal scaling} $s(\alpha) = \alpha\gamma$, i.e., the factor multiplying the target direction $Y_k$;
    \item the \emph{noise scaling}, since $\alpha U_k$ has total normalised energy $\alpha^2 e$ (by definition~\cref{eq:supp:defs}).
\end{itemize}
The global loss $\mathcal{L}$ therefore depends on $\alpha$ through these two quantities: $\mathcal{L} = \mathcal{L}\bigl(s(\alpha),\;\alpha^2 e\bigr)$.

\paragraph{Step 2. Choose the expansion point.}
The loss is minimised when the signal has optimal strength $s^\star$ and the off-target interference disappears:
\begin{equation}
\mathcal{L}_{\min} \;=\; \mathcal{L}(s^\star,\; 0).
\label{eq:supp:d5-lmin}
\end{equation}
Therefore, we perform a second-order Taylor expansion of $\mathcal{L}$ around this point $(s^\star, 0)$.

\paragraph{Step 3. First-order and cross terms vanish.}
Because $(s^\star, 0)$ is the minimiser of $\mathcal{L}$ with respect to both arguments,
\begin{equation}
\frac{\partial \mathcal{L}}{\partial s}\bigg|_{(s^\star,\,0)} = 0, \qquad \frac{\partial \mathcal{L}}{\partial n}\bigg|_{(s^\star,\,0)} = 0,
\label{eq:supp:d5-grad}
\end{equation}
where $n = \alpha\sqrt{e}$ so that $n^2 = \alpha^2e$. Furthermore, $\frac{\partial^2 \mathcal{L}}{\partial s\,\partial n}$ disappears at the expansion point, because the signal direction $Y_k$ and the noise direction $U_k$ are orthogonal by construction (condition~\cref{eq:supp:orthogonality}).

\paragraph{Step 4. Write the second-order expansion.}
With the shorthand
\begin{equation}
a \;\coloneqq\; \frac{\partial^2 \mathcal{L}}{\partial s^2}\bigg|_{(s^\star,\,0)}, \qquad b \;\coloneqq\; \frac{\partial^2 \mathcal{L}}{\partial n^2}\bigg|_{(s^\star,\,0)},
\label{eq:supp:d5-curv}
\end{equation}
the Taylor expansion to second order reads
\begin{equation}
\mathcal{L}(s,\,n) \;\approx\; \mathcal{L}_{\min} + \frac{a}{2}(s - s^\star)^2 + \frac{b}{2}\,n^2.
\label{eq:supp:d5-taylor}
\end{equation}

\paragraph{Step 5. Substitute $s = \alpha\gamma$ and $n^2 = \alpha^2 e$.}
Inserting $s(\alpha) = \alpha\gamma$ and $n = \alpha^2 e$ into~\cref{eq:supp:d5-taylor}:
\begin{equation}
\mathcal{L}(\alpha) \;\approx\; \mathcal{L}_{\min} + \frac{a}{2}(\alpha\gamma - s^\star)^2 + \frac{b}{2}\,\alpha^2 e.
\label{eq:supp:d5-sub}
\end{equation}
Therefore
\begin{equation}
\boxed{\mathcal{L}(\alpha) \;\approx\; \mathcal{L}_{\min} + \frac{a}{2}\bigl(\alpha\gamma - s^\ast\bigr)^2 + \frac{b}{2}\,\alpha^2\,e.}
\label{eq:supp:d5-result}
\end{equation}
\hfill $\blacksquare$

\vspace{1em}
\hrule
\vspace{1em}

\subsection*{Derivation 6: $\;\alpha^\star = \dfrac{s^\star\gamma}{\gamma^2 + r\,e} = \dfrac{s^\star}{\gamma}\cdot\dfrac{1}{1 + r\,\frac{e}{\gamma^2}}$}

\paragraph{Goal.} Minimise the quadratic approximation of $\mathcal{L}(\alpha)$ from Derivation~5 with respect to $\alpha$ and derive the closed-form expression for $\alpha^\star$ stated in \cref*{eq:theory:downstream:alpha_star}.

\paragraph{Step 1. Write the objective.}
From Derivation~5, we need to minimise
\begin{equation}
\mathcal{L}(\alpha) \;=\; \mathcal{L}_{\min} + \frac{a}{2}(\alpha\gamma - s^\star)^2 + \frac{b}{2}\,\alpha^2 e
\label{eq:supp:d6-objective}
\end{equation}
over $\alpha \in \mathbb{R}$. The constant $\mathcal{L}_{\min}$ does not affect the location of the minimiser, so we define
\begin{equation}
f(\alpha) \;\coloneqq\; \frac{a}{2}(\alpha\gamma - s^\star)^2 + \frac{b}{2}\,\alpha^2 e.
\label{eq:supp:d6-f}
\end{equation}

\paragraph{Step 2. Differentiate and set to zero.}
Applying the chain rule to the first term:
\begin{equation}
f'(\alpha) = a\gamma(\alpha\gamma - s^\star) + b\,\alpha\,e.
\label{eq:supp:d6-deriv}
\end{equation}
Setting $f'(\alpha^\star) = 0$:
\begin{equation}
a\gamma(\alpha^\star\gamma - s^\star) + b\,\alpha^\star e = 0.
\label{eq:supp:d6-zero}
\end{equation}

\paragraph{Step 3. Solve for $\alpha^\star$.}
Expanding~\cref{eq:supp:d6-zero}:
\begin{equation}
a\gamma^2\,\alpha^\star - a\gamma\,s^\star + b\,e\,\alpha^\star = 0.
\label{eq:supp:d6-expand}
\end{equation}
Collecting terms in $\alpha^\star$:
\begin{equation}
\alpha^\star\!\left(a\gamma^2 + b\,e\right) = a\gamma\,s^\star.
\label{eq:supp:d6-collect}
\end{equation}
Since $a > 0$ (positive curvature at a minimum) and $\gamma^2 + \frac{b}{a}e > 0$, we can divide:
\begin{equation}
\alpha^\star = \frac{a\gamma\,s^\star}{a\gamma^2 + b\,e} = \frac{s^\star\gamma}{\gamma^2 + \frac{b}{a}\,e}.
\label{eq:supp:d6-solve}
\end{equation}

\paragraph{Step 4. Derive the factored form.}
Factor $\gamma^2$ out of the denominator in~\cref{eq:supp:d6-solve}:
\begin{equation}
\gamma^2 + \frac{b}{a}\,e = \gamma^2\!\left(1 + \frac{b}{a}\cdot\frac{e}{\gamma^2}\right).
\label{eq:supp:d6-factor}
\end{equation}
Therefore
\begin{equation}
\alpha^\star = \frac{s^\star\,\gamma}{\gamma^2\!\left(1 + \frac{b}{a}\cdot\frac{e}{\gamma^2}\right)} = \frac{s^\star}{\gamma}\cdot\frac{1}{1 + \frac{b}{a}\cdot\frac{e}{\gamma^2}},
\label{eq:supp:d6-factored}
\end{equation}
which gives the second form:
\begin{equation}
\boxed{\alpha^\star = \frac{s^\star}{\gamma}\cdot\frac{1}{1 + \dfrac{b}{a}\cdot\dfrac{e}{\gamma^2}}.}
\label{eq:supp:d6-alpha-factored}
\end{equation}

\paragraph{Step 5. Verify that $\alpha^\star$ is a minimum.}
The second derivative of $f$ is
\begin{equation}
f''(\alpha) = a\gamma^2 + b\,e.
\label{eq:supp:d6-second-deriv}
\end{equation}
Since $a > 0$, $b \geq 0$, $\gamma^2 > 0$ (the router retains nonzero signal; $\gamma = 0$ would mean no task information survives), and $e \geq 0$, we have $f''(\alpha) > 0$ for all $\alpha$. Hence $\mathcal{L}(\alpha)$ is strictly convex in $\alpha$ and $\alpha^\star$ is the unique global minimiser.
\hfill $\blacksquare$

\vspace{1em}
\hrule
\vspace{1em}

\section{Additional Ablations}
\label{sec:supp:ablations}

\subsection{Comparison to PICO and LiNeS}
\label{sec:supp:ablations:picolines}

We propose to rescale merged adapters by a factor $\alpha$, which can be estimated by $1/\gamma$ (\cref*{sec:theory}). However, PICO \cite{tang2026crowded} and LiNeS \cite{wang2025lines} propose different rescaling methods for merged weight updates.
PICO proposes to rescale the Frobenius norm of the merged adapter to match the average Frobenius norm of the individual task adapters:
\begin{equation}
    \Delta W_{\text{PICO}} = \alpha_{\text{PICO}} \Delta W_{\mathrm{merged}}
\quad \mathrm{with} \quad
\alpha_{\text{PICO}} = \frac{1}{\|\Delta W_{\mathrm{PICO}}\|_F}
\cdot
\frac{1}{K}\sum_{k=1}^K \|\Delta W_k\|_F,
\end{equation}
where $\Delta W_k$ is the weight update of task $k$.
LiNeS proposes to apply a different scaling factor depending on the layer, where lower layers are scaled less and upper layers are scaled more:
\begin{equation}
    \Delta W_{\text{LiNeS}}^{(\ell)} = \alpha_{\text{LiNeS}}^{(\ell)}W_{\text{merged}}^{(\ell)} \quad\text{with}\quad \alpha_{\text{LiNeS}}^{(\ell)} = \xi + \beta\frac{\ell - 1}{L-1},
\end{equation}
where $\ell$ is the layer index (starting from 1) and $L$ the total number of layers. $\beta$ is a tunable hyperparameters, and for $\xi$ we follow the LiNeS reference implementation and estimate it as
\begin{equation}
    \xi = \frac{\|\sum_k \Delta W_k\|_1}{K \cdot \|\Delta_{\text{merged}}\|_1}.
\end{equation}
Here $\|\cdot\|_1$ denotes the entrywise $\ell_1$ norm. $\beta$ is selected as the best value from $\beta\in\{0.1, \ldots, 1.5\}$ which corresponds to the TIES grid proposed in {\NoHyper\cite{wang2025lines}}.

Results are in \cref{tab:supp:picolines}. Scaling the plain SSR-merged adapter with either LiNeS or PICO hurts performance throughout, and performance decreases fast as $K$ increases. This shows that our global scaling is fundamentally different from PICO and LiNeS scaling, and they are not directly applicable in our setting.

\begin{table*}[t]
\centering
%\footnotesize
\setlength{\tabcolsep}{3pt}
%\resizebox{\textwidth}{!}{%
\begin{tabular}{l ccccccccc ccccccccc}
\toprule
\multirow{3.5}{*}{\textbf{Method}}
  & \multicolumn{8}{c}{\textbf{FLUX.1-dev}} && \multicolumn{8}{c}{\textbf{SD3.5-Medium}} \\
\cmidrule(lr){2-9} \cmidrule(lr){11-18}
  & \multicolumn{2}{c}{$3$} & \multicolumn{2}{c}{$9$} & \multicolumn{2}{c}{$15$} & \multicolumn{2}{c}{$30$} 
  && \multicolumn{2}{c}{$3$} & \multicolumn{2}{c}{$9$} & \multicolumn{2}{c}{$15$} & \multicolumn{2}{c}{$30$} \\
\cmidrule(lr){2-3} \cmidrule(lr){4-5} \cmidrule(lr){6-7} \cmidrule(lr){8-9}
\cmidrule(lr){11-12} \cmidrule(lr){13-14} \cmidrule(lr){15-16} \cmidrule(lr){17-18}
  & D & I & D & I & D & I & D & I && D & I & D & I & D & I & D & I \\
\midrule
SSR\ ($\alpha=1$)             & .680 & .816 & .649 & .798 & .628 & .792 & .607 & .788 && .628 & .784 & .589 & .768 & .574 & .759 & .569 & .757 \\
SSR\,+\,PICO                  & .558 & .753 & .514 & .742 & .500 & .740 & .490 & .737 && .524 & .741 & .472 & .715 & .463 & .711 & .454 & .707 \\
SSR\,+\,LiNeS                 & .669   & .812   & .654   & .802   & .623   & .792   & .561 & .771 && .662   & .792   & .634   & .773   & .614   & .762   & .573 & .747 \\
\midrule
SSR\,+\,$1/\gamma$ (ours)     & \underline{.707} & \underline{.825} & \underline{.697} & \underline{.821} & \textbf{.691} & \textbf{.820} & \underline{.674} & \underline{.815} && \underline{.664} & \underline{.797} & \underline{.646} & \underline{.788} & \underline{.640} & \underline{.781} & \underline{.626} & \textbf{.774} \\
SSR\,+\,line search (ours)    & \textbf{.716} & \textbf{.831} & \textbf{.711} & \textbf{.828} & \underline{.681} & \underline{.819} & \textbf{.687} & \textbf{.819} && \textbf{.680} & \textbf{.799} & \textbf{.668} & \textbf{.789} & \textbf{.649} & \textbf{.783} & \textbf{.635} & \underline{.773} \\
\midrule
{\textit{Upper Bound}}           & \UB{.723} & \UB{.833} & \UB{.723} & \UB{.833} & \UB{.723} & \UB{.833} & \UB{.723} & \UB{.833} && \UB{.671} & \UB{.804} & \UB{.671} & \UB{.804} & \UB{.671} & \UB{.804} & \UB{.671} & \UB{.804} \\
\bottomrule
\end{tabular}%
%}
\caption{\textbf{Comparison against existing rescaling strategies (PICO and LiNeS)} across various pool sizes $K \in \{3, 9, 15, 30\}$ on DreamBooth. We report subject fidelity using D (DINOv2-B) and I (CLIP-I ViT-L/14). Each cell represents the mean over all 30 target subjects. \textbf{Bold} indicates the best performance among compared methods, and \underline{underline} denotes the second best.}
\label{tab:supp:picolines}
\end{table*}

\subsection{Ablation of the Calibration Data}
\label{sec:supp:calibrationdata}

{\noindent \textbf{Setting.}}
Wei et al.\ find that for plain SSR-Merging, adding more calibration data such as additional prompt templates and data from the full diffusion trajectory (i.e., all diffusion steps) does not improve performance. However, this is counterintuitive, as adapters are applied at all diffusion steps, and activations differ significantly between them. Moreover, empirically, we find the best performance of SSR + $1/\gamma$ and SSR + Line Search when including activations from the full diffusion trajectory in the calibration data.

To ablate the calibration data, we systematically compare three methods (SSR, SSR + $1/\gamma$, and SSR + Line Search) on four configurations: (One step, Trajectory) $\times$ (generic prompt, multiple prompts). One step vs.\ Trajectory compares using only the first diffusion step to collect activations for training the router $R$ with using activations from all diffusion steps. Generic prompt vs.\ multiple prompts compares using a prompt that mentions the subject (e.g., \enquote{dachshund}), but no scene, evaluated on four different seeds for sampling diffusion noise, with four prompts that mention the subject and in addition a scene (e.g., \enquote{dachshund in a garden}).

\begin{figure*}
\centering
\includegraphics[width=\linewidth]{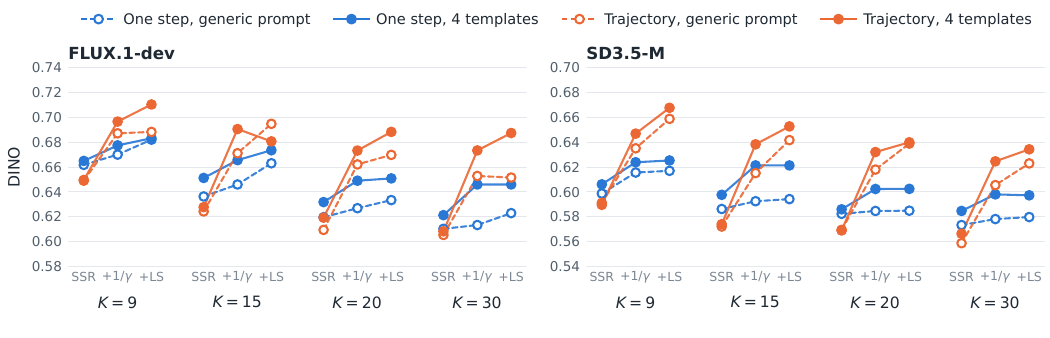}
\caption{DINO similarity on FLUX.1-dev and SD3.5-M for SSR, SSR\,$+$\,$1/\gamma$, and SSR\,$+$\,line search (+LS). We compare calibration activations collected from a single diffusion step versus the full trajectory, paired with either a generic prompt or four prompt templates.}
\label{fig:ablations:calibrationdata}
\end{figure*}

\mypara{Results}
In \cref{fig:ablations:calibrationdata}, we show the DINO metric for FLUX and SD3.5-M across $K = 9, 15, 20, 30$. The results clearly show that more diverse calibration data helps little or hurts for plain SSR, as Wei et al.\ observed. However, our scaling effectively unlocks significant gains from better calibration data. Only when scaling merged adapters with $1/\gamma$ or $\alpha$ from Line Search can we make use of more in-distribution data. Therefore, our method achieves proper scaling behavior, which we expect from training-based merging methods. Furthermore, we find that gains come from both full-trajectory activations and more in-distribution prompts. In isolation, both yield similar gains, with full-trajectory activations improving performance slightly better. Combined, they yield further gains.

\subsection{Runtime of Activation-Based Merging Methods}
\label{sec:supp:runtime}

To compare the computational cost of the activation-based merging methods, we measure the wall time and the number of floating-point operations (FLOPs) of a single merge on the DreamBooth task. Each merge combines the K adapters of one configuration ($K = 3, 9, 15, 30$ on SD3.5-Medium; $K = 30$ on FLUX.1-dev). All methods use the same calibration data, i.e., we evaluate a data-matched scenario: four prompt templates, each with the full 28-step sampling trajectory. Every merge starts from scratch, without cached features or stored statistics, and runs on a single GH200 GPU. We exclude the time to load the model and the adapters. For SSR, the cost includes collecting the subspace statistics of the K adapters. For SSR with line search, it also includes generating the validation images of all $K$ subjects and scoring them with DINO and CLIP, for each of the seven values of $\alpha$ from which we select the best one. On SD3.5-Medium, all methods pass over the calibration data once per adapter. On FLUX, the Gram matrices of RegMean and Mix-of-Show do not fit in GPU memory next to the model, so we compute them in chunks over replayed trajectories. We count FLOPs with PyTorch's FLOP counter and add analytic counts for linear solves. For IterIS, we time two of its ten iterations and extrapolate to the full run due to the excessive runtime. On FLUX, IterIS is not possible to run with 30 adapters on a GH200 server, so results in \cref*{tab:dreambooth} were obtained on B300 GPUs.

Results are in \cref{tab:supp:merge_cost}. Estimating $\gamma$ add only small overhead on top of SSR, while line search is slower by factor $2\times$ to $3\times$ because it needs to evaluate tasks multiple times. SSR from the start is slower than Regmean abd Mix-of-Show, but not computationally more expensive in terms of FLOPS. However, it scales to larger models with higher activation dimensions, where RegMean and Mix-of-Show break down even on recent hardware. IterIS is very costly throughout and requires state-of-the-art hardware to run on FLUX.

\begin{table*}[t]
\centering
\footnotesize
\setlength{\tabcolsep}{4pt}
\begin{tabular}{l cc cc cc cc cc}
\toprule
\multirow{3.5}{*}{\textbf{Method}} & \multicolumn{2}{c}{\textbf{FLUX.1-dev}} & \multicolumn{8}{c}{\textbf{SD3.5-Medium}} \\
\cmidrule(lr){2-3} \cmidrule(lr){4-11} 
& \multicolumn{2}{c}{$K{=}30$} & \multicolumn{2}{c}{$K{=}3$} & \multicolumn{2}{c}{$K{=}9$} & \multicolumn{2}{c}{$K{=}15$} & \multicolumn{2}{c}{$K{=}30$} \\
\cmidrule(lr){2-3} \cmidrule(lr){4-5} \cmidrule(lr){6-7} \cmidrule(lr){8-9} \cmidrule(lr){10-11}
& min & PFLOP & min & PFLOP & min & PFLOP & min & PFLOP & min & PFLOP \\
\midrule
% ---- activation-based merging
RegMean\calib       & 144 & 478 & 0.5 & 1.9 & 1.5 & 5.8 & 2.4 & 9.6 & 4.4 & 19 \\
Mix-of-Show\calib   & 170 & 485 & 1.0 & 2.1 & 2.0 & 5.9 & 2.7 & 9.8 & 4.9 & 19 \\
IterIS\calib        & \multicolumn{2}{c}{OOM} & 9.9 & 43 & 27\rlap{$^\dagger$} & 128 & 46\rlap{$^\dagger$} & 213 & 94\rlap{$^\dagger$} & 426 \\
\midrule
% ---- SSR
SSR\calib           & 11 & 84 & 0.5 & 1.7 & 1.7 & 5.4 & 2.9 & 9.4 & 6.0 & 21 \\
SSR\,+\,$1/\gamma$\calib (ours) & 13 & 84 & 0.6 & 1.7 & 1.9 & 5.4 & 2.9 & 9.4 & 6.4 & 21 \\
SSR\,+\,line search\calib (ours) & 43 & 718\rlap{$^\ddagger$} & 1.6 & 14 & 5.0 & 45 & 8.6 & 79 & 17 & 179 \\
\bottomrule
\end{tabular}
\caption{Wall time in minutes on one GH200 GPU and compute in PFLOP for relevant methods. All methods use the same calibration data (4 prompt templates $\times$ the full 28-step sampling trajectory per adapter). SSR methods include computing the subspace statistics of the $K$ adapters and line search additionally generates and scores (DINO, CLIP) the validation images of all $K$ subjects. $^\dagger$2 of 10 IterIS iterations measured and extrapolated (at $K{=}3$). $^\ddagger$Validation FLOPs extrapolated from one subject to all $K$. OOM: out of GPU memory.}
\label{tab:supp:merge_cost}
\end{table*}

\subsection{Comparing Heuristic and Line Search Scaling Factors}
\label{sec:supp:gammaalpha}

\cref{tab:supp:gamma_alpha} compares the $1/\gamma$ heuristic with the scaling factor $\alpha$ selected by line search. Across all settings, both factors exceed $1$, confirming that compensating for SSR shrinkage improves merged performance. Whether $1/\gamma$ over- or underestimates $\alpha$ depends on the task and layer group.

On DreamBooth, shrinkage is stronger in the image stream ($\gamma \in [0.51, 0.60]$) than in the text stream ($\gamma \in [0.68, 0.83]$). The resulting image-stream gains ($1/\gamma \in [1.67, 1.97]$) match the global line-search $\alpha$ ($1.75$-$2.00$) much more closely than the smaller text-stream gains ($1.21$--$1.47$), which indicates that correcting the image stream is overall more important. Among settings with a single pooled gain, $1/\gamma$ underestimates $\alpha$ on FLUX.1-dev style transfer ($1.31$ vs.\ $1.78$) and Llama generation ($1.40$ vs.\ $1.77$), where $K=3$ adapters induce little interference ($\gamma = 0.72$). Conversely, $1/\gamma$ exceeds $\alpha$ on image classification, Llama NLI and combined tasks, and SD3.5-M style transfer. This pattern aligns with our theoretical optimum $\alpha^\star$: when shrinkage is stronger and off-target interference is larger, amplifying the merged residual becomes costly, pulling the optimal scaling factor below $1/\gamma$.

\begin{table*}[t]
\centering
\setlength{\tabcolsep}{3.2pt}
% ---------------- (a) DreamBooth: per layer group ----------------
\begin{tabular}{l ccc ccc ccc cc cc cc}
\toprule
& \multicolumn{15}{c}{\textbf{DreamBooth}} \\
\cmidrule(lr){2-16}
& \multicolumn{9}{c}{FLUX.1-dev} & \multicolumn{6}{c}{SD3.5-M} \\
\cmidrule(lr){2-10} \cmidrule(lr){11-16}
& \multicolumn{3}{c}{$K{=}9$} & \multicolumn{3}{c}{$K{=}20$} & \multicolumn{3}{c}{$K{=}30$}
& \multicolumn{2}{c}{$K{=}9$} & \multicolumn{2}{c}{$K{=}20$} & \multicolumn{2}{c}{$K{=}30$} \\
\cmidrule(lr){2-4} \cmidrule(lr){5-7} \cmidrule(lr){8-10} \cmidrule(lr){11-12} \cmidrule(lr){13-14} \cmidrule(lr){15-16}
& txt & img & sgl & txt & img & sgl & txt & img & sgl & txt & img & txt & img & txt & img \\
\midrule
$\gamma$               & .83  & .60  & .69  & .82  & .55  & .64  & .81  & .54  & .62  & .72  & .57  & .69  & .52  & .68  & .51  \\
$1/\gamma$             & 1.21 & 1.67 & 1.45 & 1.22 & 1.82 & 1.57 & 1.23 & 1.87 & 1.61 & 1.40 & 1.76 & 1.46 & 1.92 & 1.47 & 1.97 \\
$\alpha$ (line search) & \multicolumn{3}{c}{1.75} & \multicolumn{3}{c}{1.75} & \multicolumn{3}{c}{1.75}
& \multicolumn{2}{c}{2.00} & \multicolumn{2}{c}{1.75} & \multicolumn{2}{c}{2.00} \\
\bottomrule
\end{tabular}

\vspace{0.8em}

% ---------------- (b) other tasks: pooled over all layers ----------------
\begin{tabular}{l cc cc ccc ccc ccc}
\toprule
& \multicolumn{4}{c}{\textbf{Style Transfer}} & \multicolumn{6}{c}{\textbf{Image Classification}} & \multicolumn{3}{c}{\textbf{Llama-3.1-8B}} \\
\cmidrule(lr){2-5} \cmidrule(lr){6-11} \cmidrule(lr){12-14}
& \multicolumn{2}{c}{FLUX.1-dev} & \multicolumn{2}{c}{SD3.5-M} & \multicolumn{3}{c}{ViT-B/32} & \multicolumn{3}{c}{ViT-L/14} & NLI & gen & comb. \\
\cmidrule(lr){2-3} \cmidrule(lr){4-5} \cmidrule(lr){6-8} \cmidrule(lr){9-11} \cmidrule(lr){12-12} \cmidrule(lr){13-13} \cmidrule(lr){14-14}
$K$                    & 9    & 20   & 9    & 20   & 8    & 14   & 20   & 8    & 14   & 20   & 6    & 3    & 9    \\
\midrule
$\gamma$               & .77  & .76  & .66  & .64  & .61  & .55  & .52  & .44  & .39  & .36  & .50  & .72  & .55  \\
$1/\gamma$             & 1.30 & 1.31 & 1.52 & 1.56 & 1.65 & 1.82 & 1.94 & 2.28 & 2.60 & 2.80 & 2.00 & 1.40 & 1.83 \\
$\alpha$ (line search) & 1.86 & 1.78 & 2.57 & 1.35 & 1.59 & 1.71 & 1.75 & 2.11 & 2.38 & 2.42 & 1.66 & 1.77 & 1.68 \\
\bottomrule
\end{tabular}
\caption{Measured shrinkage $\gamma$ of SSR, the closed-form gain $1/\gamma$ applied by SSR\,+\,$1/\gamma$, and the gain $\alpha$ selected by line search on validation data. On DreamBooth, SSR\,+\,$1/\gamma$ applies one gain per layer group, so we report $\gamma$ and $1/\gamma$ separately for the text-stream (txt) and image-stream (img) layers of the dual-stream blocks and, for FLUX.1-dev, the single-stream blocks (sgl). Line Search selects one global $\alpha$.}
\label{tab:supp:gamma_alpha}
\end{table*}

\subsection{Comparing Per-Layer and Global Rescaling Factors}
\label{sec:supp:perlayer}

For our experiments, we use a global scaling factor instead of per-layer factors. For image generation when scaling with $1/\gamma$, we apply different scaling factors per layer group, for example the text-only stream, image-only stream, and shared streams in Flux. For line search, this is necessary, because simultaneously optimizing multiple factors is computationally too costly. For the $1/\gamma$ heuristic, we observe empirically better results when pooling $\gamma$ across layers on SSR, but better results when rescaling layers with their individually $\gamma$ values on RegMean, see \cref{tab:supp:gamma_layer}. On both SSR and RegMean, the optimal configuration depends on the subject. We think that in principle, per layer scaling is optimal, but $1/\gamma$ is a noisy heuristic that can be stabilized by pooling.

\begin{table}[t]
\centering
\small
\setlength{\tabcolsep}{4pt}
\begin{tabular}{l cccc c cccc}
\toprule
& \multicolumn{4}{c}{\textbf{SSR}} && \multicolumn{4}{c}{\textbf{RegMean}} \\
\cmidrule(lr){2-5} \cmidrule(lr){7-10}
$K$ & 9 & 15 & 20 & 30 && 9 & 15 & 20 & 30 \\
\midrule
No rescaling           & .590 & .574 & .569 & .566 && .619 & .587 & .572 & .553 \\
Group $1/\gamma$       & \textbf{.647} & \textbf{.638} & \textbf{.632} & \textbf{.625} && .644 & .635 & .623 & .600 \\
Per-layer $1/\gamma$   & .645 & .625 & .610 & .592 && \textbf{.649} & \textbf{.636} & \textbf{.629} & \textbf{.611} \\
\midrule
$\Delta$ DINO          & $-$.002 & $-$.013 & $-$.022 & $-$.033 && $+$.006 & $+$.001 & $+$.006 & $+$.011 \\
Per-layer better       & 13/30 & 11/30 & 12/30 & 7/30 && 14/30 & 18/30 & 16/30 & 19/30 \\
\bottomrule
\end{tabular}
\caption{Per-layer vs.\ layer-group $1/\gamma$. DreamBooth on SD3.5-Medium, scores are mean DINO over 30 subjects per $K$. Per-layer estimates $\gamma$ for each of the 191 adapted layers. $\Delta$ means per-layer score $-$ group score. Bold: best rescaling per column.}
\label{tab:supp:gamma_layer}
\end{table}

\section{Additional Experimental Details}
\label{sec:supp:additionaldetails}

\subsection{Estimating the Rescaling Factor from Calibration Data}
\label{sec:supp:gamma}

To compute the heuristic rescaling factor $\hat{\alpha} = 1/\gamma$, we estimate $\gamma$ directly from the calibration data. Computing $\gamma$ on the exact same activations used to solve the least-squares objective would introduce an optimistic in-sample bias (underestimating shrinkage). Therefore, we estimate $\gamma$ via cross-validation over the calibration set.

\mypara{Out-of-Sample Split}
We partition the calibration inputs into $P$ disjoint splits: two halves ($P=2$) for classification and LLM tasks, and one split per prompt template ($P=4$) for diffusion tasks. For each fold $p \in \{1, \dots, P\}$, the merge is fitted on all remaining splits $\{-p\}$ and evaluated on split $p$. For a given layer $\ell$ and task $k$ with $m_k$ calibration tokens in split $p$, we compute the sample inner product and target norm:
\begin{equation}\label{eq:gamma:numden}
    \mathrm{num}_{k,\ell}^{(p)} = \frac{1}{m_k} \langle \hat{Y}_k^{(-p)}, Y_k \rangle_F, \qquad
    \mathrm{den}_{k,\ell}^{(p)} = \frac{1}{m_k} \|Y_k\|_F^2,
\end{equation}
where $Y_k = B_k A_k X_k$ is the standalone adapter output on split $p$, and $\hat{Y}_k^{(-p)}$ is the prediction produced by the merge fitted on all splits except $p$.

\mypara{Formulation for RegMean and Mix-of-Show}
In weight-space merging methods such as RegMean and Mix-of-Show, the merged update is a full matrix $\Delta W \in \mathbb{R}^{d \times d}$, yielding predictions $\hat{Y}_k^{(-p)} = \Delta W^{(-p)} X_k$. Letting $\Sigma_k = \frac{1}{m_k} X_k X_k^\top \in \mathbb{R}^{d \times d}$ denote the sample covariance of the input activations on split $p$, \cref{eq:gamma:numden} evaluates to:
\begin{align}
    \mathrm{num}_{k,\ell}^{(p)} &= \mathrm{tr}\big( B_k^\top \Delta W^{(-p)} \Sigma_k A_k^\top \big), \label{eq:gamma:regmean:num} \\
    \mathrm{den}_{k,\ell}^{(p)} &= \mathrm{tr}\big( B_k^\top B_k A_k \Sigma_k A_k^\top \big). \label{eq:gamma:regmean:den}
\end{align}
Evaluating \cref{eq:gamma:regmean:num,eq:gamma:regmean:den} requires computing and storing full $d \times d$ activation Gram matrices $\Sigma_k$ for every task, layer, and split, followed by refitting the full matrix $\Delta W^{(-p)}$.

\mypara{Formulation for SSR-Merge}
In contrast, SSR preserves the factorized LoRA structure by routing through the bottleneck activations $Z_k = A_{\text{comb}} X_k \in \mathbb{R}^{Nr \times m_k}$. The target signal for task $k$ is routed directly into its own up-projection $B_k$ via the $k$-th block row of the router, $R_k \in \mathbb{R}^{r \times Nr}$, so that $\hat{Y}_k = B_k R_k Z_k$. 

Let $M_k = \frac{1}{m_k} Z_k Z_k^\top \in \mathbb{R}^{Nr \times Nr}$ denote the bottleneck second-moment matrix on split $p$. Because the $k$-th block of $Z_k$ is $A_k X_k$, the cross-covariance between the concatenated bottleneck representations $Z_k$ and the adapter's own bottleneck representation $A_k X_k$ is given by the $k$-th block column of $M_k$, denoted $M_k^{(:,k)} \in \mathbb{R}^{Nr \times r}$. Similarly, the auto-covariance of $A_k X_k$ is the $k$-th diagonal block $M_k^{(k,k)} \in \mathbb{R}^{r \times r}$. Substituting these identities into \cref{eq:gamma:numden} yields:
\begin{align}
    \mathrm{num}_{k,\ell}^{(p)} &= \mathrm{tr}\big( B_k^\top B_k R_k^{(-p)} M_k^{(:,k)} \big), \label{eq:gamma:ssr:num} \\
    \mathrm{den}_{k,\ell}^{(p)} &= \mathrm{tr}\big( B_k^\top B_k M_k^{(k,k)} \big). \label{eq:gamma:ssr:den}
\end{align}
Notice that \cref{eq:gamma:ssr:num,eq:gamma:ssr:den} depend only on the $Nr \times Nr$ bottleneck moments $M_k$ (which SSR already collects during merging) and the small $r \times r$ projection matrices $B_k^\top B_k$. No $d \times d$ covariance matrix is ever formed or stored, meaning that cross-validating $R^{(-p)}$ and computing $\gamma$ introduces no additional computational or memory overhead.

\mypara{Aggregation and Final Rescaling}
To obtain the final shrinkage factor $\gamma$, we aggregate the per-fold statistics: For Classification and LLM models, we pool the terms across all tasks, layers, and folds:
\begin{equation}
    \gamma = \frac{\sum_{p, \ell, k} \mathrm{num}_{k,\ell}^{(p)}}{\sum_{p, \ell, k} \mathrm{den}_{k,\ell}^{(p)}}.
\end{equation}
For diffusion models, we average the per-fold ratios over the $P$ prompt folds. Furthermore, modern diffusion architectures consist of heterogeneous block types (e.g., text, image, and single-stream transformer blocks in FLUX). We compute a separate $\gamma$ for each block type by restricting the sum over $\ell$ to layers of that type.
Once $\gamma$ is computed, the final router $R^{\star}$ (or weight matrix $\Delta W^{\star}$) is fitted on all calibration data and rescaled by $1/\gamma$.

\subsection{Hyperarameter Tuning of the Baselines}
\label{sec:supp:hyperparameters}

For our results in \cref*{tab:dreambooth} we tuned all baselines following recommendations from their respective reference implementations or papers, where available. RegMean, Mix-of-Show, and IterIS did not provide any recommendations, so we used the constant settings from their official implementations. The tuned hyperparameters of all methods are in \cref{tab:supp:hyperparameter-grid}. Hyperparameters are tuned individually for each number of merged adapters, i.e., we get one configuration per method and $K$. Selected values are in \cref{tab:supp:selected-hyperparameters} for some methods, optimal hyperparameters differ substantially for different $K$ which shows the need for hyperparameter optimization.

All methods use the same validation set which includes 10 subjects from the original SSR-Merge subjects. Each subject is rendered with two validation
prompts (\enquote{a sks \{class\} on a cobblestone street}, \enquote{a sks \{class\} with a city in the background}) and two seeds, i.e.\ 4 images per subject. None of these prompts is used at test time. We then score the hyperparameter with mean DINOv2 similarity over the 10 validation subjects.
Methods that use activations (RegMean, Mix-of-Show, IterIS, SSR) use the same calibration data, i.e., four prompt templates, full 28-step trajectories of the respective individual LoRA, seed 0.

\begin{table*}[t]
\centering
\footnotesize
\setlength{\tabcolsep}{4pt}
\begin{tabular}{l l l p{5.8cm}}
\toprule
\textbf{Method} & \textbf{Tuned} & \textbf{Grid (validation)} & \textbf{Fixed settings} \\
\midrule
Task Arithmetic & $\lambda$ & $\{0.1, 0.2, \dots, 1.0\}$ & $\Delta W = \lambda \sum_k B_k A_k$ \\
TIES            & $\lambda$ & $\{0.8, 0.9, \dots, 1.8\}$ & top-$20\%$ trimming per task, sign election, disjoint mean \\
DARE            & $p$       & $\{0.1, 0.3, 0.5, 0.7, 0.9\}$ & on Task Arithmetic; $\lambda$ from TA; rescale $1/(1-p)$ \\
RobustMerge     & --        & -- & top-$20\%$ magnitude mask per task, rank-wise scaling, fuse $=2.0$ \\
KnOTS           & $\alpha$, top-$k$ & \makecell[l]{(i) $\alpha \in \{0.1, \dots, 1.0\}$ at top-$k{=}30\%$ \\ (ii) top-$k \in \{10, 20, \dots, 100\}\%$ at selected $\alpha$} & Applied on TIES \\
Core Space      & Iso, $\alpha$ & $\{\text{off}, \text{on}\} \times \{0.1, 0.2, \dots, 3.0\}$ & TSV merge in core space \\
Iso-C           & $\alpha$  & $\{0.1, 0.2, \dots, 3.0\}$ & applied in core-space coordinates  \\
\midrule
RegMean         & --        & -- & off-diagonal Gram scaling $0.9$ \\
Mix-of-Show     & --        & -- & gradient fusion; L-BFGS, lr $1$, history $25$, strong Wolfe, $50$ steps from $W_0$ \\
IterIS          & --        & -- & $10$ iterations, $\alpha_1 = \alpha_2 = 10^{-7}$ \\
\midrule
SSR             & --        & -- & -- \\
SSR\,+\,$1/\gamma$ (ours) & -- & -- & $\alpha = 1/\gamma$ per layer group, leave-one-template-out over the 4 calibration templates \\
SSR\,+\,line search (ours) & $\alpha$ & $\{1.0, 1.25, 1.5, 1.75, 2.0\}$ & global scaling \\
\bottomrule
\end{tabular}
\caption{Search spaces of all methods in \cref*{tab:dreambooth}. Numbers in parentheses are the number of validation runs per $K$ (each over 10 configurations). Methods without a grid are run once with the listed constants.}
\label{tab:supp:hyperparameter-grid}
\end{table*}

\begin{table*}[t]
\centering
\footnotesize
\setlength{\tabcolsep}{4pt}
\begin{tabular}{l l cccccc cccccc}
\toprule
& & \multicolumn{6}{c}{\textbf{FLUX.1-dev}} & \multicolumn{6}{c}{\textbf{SD3.5-Medium}} \\
\cmidrule(lr){3-8} \cmidrule(lr){9-14}
\textbf{Method} & \textbf{HP} & $3$ & $5$ & $9$ & $15$ & $20$ & $30$ & $3$ & $5$ & $9$ & $15$ & $20$ & $30$ \\
\midrule
Task Arithmetic & $\lambda$ & 1.0 & 0.9 & 0.8 & 0.3 & 0.4 & 0.2 & 0.9 & 1.0 & 0.6 & 0.4 & 0.3 & 0.2 \\
TIES            & $\lambda$ & 1.8 & 1.6 & 1.8 & 1.5 & 1.8 & 1.1 & 1.7 & 1.8 & 1.4 & 1.2 & 1.2 & 1.1 \\
DARE            & $p$       & 0.5 & 0.3 & 0.1 & 0.5 & 0.7 & 0.5 & 0.3 & 0.1 & 0.9 & 0.3 & 0.1 & 0.1 \\
\multirow{2}{*}{KnOTS}      & $\alpha$       & 1.0 & 1.0 & 1.0 & 0.9 & 0.8 & 1.0 & 1.0 & 1.0 & 1.0 & 1.0 & 1.0 & 1.0 \\
                            & top-$k$ (\%)   & 10  & 10  & 10  & 30  & 10  & 10  & 10  & 10  & 10  & 10  & 10  & 10  \\
\multirow{2}{*}{Core Space} & Iso            & off & off & off & off & off & on  & off & off & off & off & off & off \\
                            & $\alpha$       & 1.3 & 1.4 & 1.2 & 1.1 & 1.1 & 2.6 & 1.1 & 1.3 & 1.1 & 1.0 & 0.9 & 0.8 \\
Iso-C           & $\alpha$  & 3.0 & 3.0 & 3.0 & 2.9 & 2.9 & 2.6 & 3.0 & 3.0 & 2.9 & 2.9 & 2.8 & 3.0 \\
\midrule
SSR\,+\,line search (ours) & $\alpha$ & 1.50 & 1.75 & 1.75 & 2.00 & 2.00 & 1.75 & 2.00 & 2.00 & 2.00 & 2.00 & 1.75 & 2.00 \\
\bottomrule
\end{tabular}
\caption{Selected hyperparameter values in \cref*{tab:dreambooth} for all methods.}
\label{tab:supp:selected-hyperparameters}
\end{table*}

\subsection{WikiArt Style Transfer Details}
\label{sec:supp:wikiart}

\begin{figure}[t]
\centering
\includegraphics[width=\linewidth]{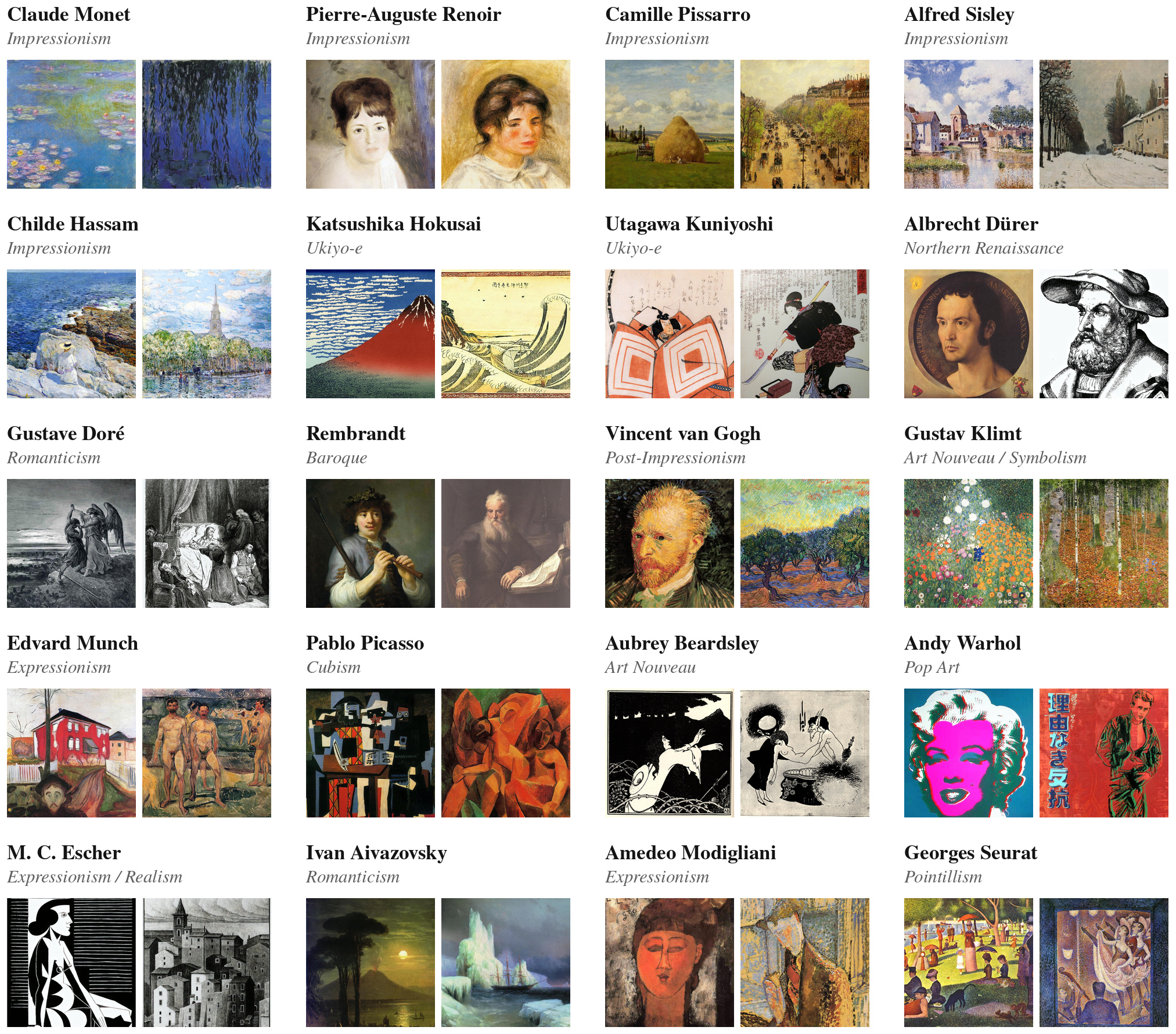}
\caption{Example reference images for all 20 artists included in our WikiArt Style Transfer experiment.}
\label{fig:supp:wikiart:data}
\end{figure}

Our WikiArt style transfer task contains works by 20 different artists belonging to different stylistic groups. Specifically, our selection includes 5 impressionists (Claude Monet, Pierre-Auguste Renoir, Camille Pissarro, Alfred Sisley, and Childe Hassam) and two authors of Ukiyo-E (Katsushika Hokusai and Utagawa Kuniyoshi). For each artist, we only sampled works labeled as the respective assigned style. An overview of all included artists with example artworks is in \cref{fig:supp:wikiart:data}.

\subsection{Concept Erasure Details}
\label{sec:supp:erasure}

\mypara{Erasers}
We train one eraser per CIFAR-10 class on Stable Diffusion v1.4~\cite{rombach2022highresolution} with the official MACE implementation and its object-erasure configuration~\cite{lu2024mace} (closed-form cross-attention refinement, rank-1 LoRA, 50 steps, learning rate $10^{-5}$, single-concept fusion). MACE only changes the key and value projections of cross-attention. We factorize each change $\Delta W = W_{\text{MACE}} - W_0$ exactly by an SVD, $\Delta W = (U\Sigma)V^\top$, so the resulting LoRA reproduces the MACE model up to floating-point precision and all methods merge the same adapters.

\mypara{Configurations}
For $K{=}3$ we use three disjoint triples (\eg $\{$bird, cat, frog$\}$), for $K{=}5$ the two halves of the ten classes, and for $K{=}7$ five windows of seven consecutive classes in a fixed permutation. The remaining classes are preserved. All numbers are averaged over the configurations of a given $K$.

\mypara{Evaluation}
We follow MACE~\cite{lu2024mace}: prompt \enquote{a photo of the \{class\}}, DDIM with 50 steps, guidance 7.5, $512{\times}512$. A zero-shot CLIP ViT-B/32 classifier over the ten classes scores each image by the softmax probability of its class. E${}=1-\mathrm{Acc}_e$ is measured on the erased classes, P${}=\mathrm{Acc}_s$ on the preserved classes, generality $1-\mathrm{Acc}_g$ on MACE's three synonyms per erased class (\eg aircraft, plane, jet), and $H_o$ is their harmonic mean. For compute reasons we use 25 images per class and 8 per synonym instead of 200.

\mypara{Calibration and preservation term}
Each eraser is calibrated alone on its class name in four templates (\eg \enquote{a painting of a \{class\}}), none of which is used for validation or testing. The key and value projections only see CLIP text embeddings, so one denoising step per prompt suffices. Methods marked $\dagger$ add the preservation term of MACE and UCE~\cite{lu2024mace,gandikota2024unified}, which pushes the merged update towards zero on base-model activations of 30 neutral concepts (\eg chair, teapot, lighthouse) in the same templates, with the same weight $w{=}3$ for SSR, RegMean, and MACE. $\gamma$ is estimated as in \cref{sec:supp:gamma} with one fold per template.

\mypara{Baselines and hyperparameters}
We use the baselines and grids of \cref{tab:supp:hyperparameter-grid}, with a step of 0.2 for $\alpha$ in Iso-C and Core Space. As an erasure-specific baseline we add MACE's closed-form multi-LoRA fusion~\cite{lu2024mace}, i.e., RegMean without off-diagonal scaling plus the preservation term. SSR\,+\,line search scans $\alpha \in \{0.75, 0.875, \dots, 3.0\}$ and refines the best value with a quadratic fit. All hyperparameters are selected per $K$ by $H_o$ on validation prompts (\enquote{an image of the \{class\}}) with seeds disjoint from the test set (\cref{tab:supp:erasure:hp}).

\begin{table*}[t]
\centering
\footnotesize
\setlength{\tabcolsep}{3.5pt}
\begin{tabular}{l ccc ccc ccc ccc ccc}
\toprule
& \multicolumn{9}{c}{CIFAR-10 objects} & \multicolumn{6}{c}{Celebrities} \\
\cmidrule(lr){2-10}\cmidrule(lr){11-16}
\multirow{2.5}{*}{\textbf{Method}} & \multicolumn{3}{c}{$K{=}3$} & \multicolumn{3}{c}{$K{=}5$} & \multicolumn{3}{c}{$K{=}7$} & \multicolumn{3}{c}{$K{=}3$} & \multicolumn{3}{c}{$K{=}10$} \\
\cmidrule(lr){2-4} \cmidrule(lr){5-7} \cmidrule(lr){8-10} \cmidrule(lr){11-13} \cmidrule(lr){14-16}
& E & P & $H_o$ & E & P & $H_o$ & E & P & $H_o$ & E & P & $H_o$ & E & P & $H_o$ \\
\midrule
\multicolumn{16}{l}{\textit{Weight-based}} \\
Task Arithmetic & .88 & .71 & .813 & .79 & .49 & .653 & .74 & .47 & .619 & .99 & .82 & \underline{.899} & .85 & .75 & .795 \\
TIES & .82 & .80 & .815 & .78 & .56 & .693 & .73 & .57 & .678 & .93 & .85 & .887 & .62 & .70 & .654 \\
DARE & .88 & .65 & .789 & .75 & .52 & .663 & .74 & .45 & .611 & .97 & .84 & \underline{.899} & .89 & .74 & \underline{.805} \\
RobustMerge & .90 & .34 & .562 & .87 & .13 & .293 & .87 & .11 & .263 & .98 & .68 & .799 & .81 & .62 & .694 \\
KnOTS & .20 & .98 & .324 & .02 & .98 & .056 & .04 & .98 & .093 & .36 & .81 & .495 & .14 & .82 & .233 \\
Core Space & .76 & .90 & .810 & .76 & .71 & .739 & .78 & .66 & .727 & .64 & .76 & .685 & .47 & .71 & .555 \\
Iso-C & .55 & .69 & .622 & .47 & .59 & .528 & .65 & .37 & .532 & .40 & .47 & .395 & .40 & .47 & .430 \\
\midrule
\multicolumn{16}{l}{\textit{Activation-based}} \\
Mix-of-Show & .86 & .98 & \underline{.869} & .86 & .94 & .866 & .86 & .88 & .843 & 1.00 & .75 & .855 & .99 & .54 & .699 \\
IterIS & .83 & .96 & .861 & .86 & .88 & .859 & .86 & .77 & .820 & 1.00 & .79 & .880 & .99 & .57 & .726 \\
RegMean$^\dagger$ & .83 & .98 & .828 & .84 & .99 & .839 & .82 & .99 & .819 & .98 & .84 & \textbf{.906} & .90 & .78 & \textbf{.832} \\
MACE$^\dagger$ & .84 & .98 & .837 & .86 & .94 & .863 & .85 & .93 & \underline{.851} & .99 & .81 & .891 & .98 & .64 & .772 \\
SSR$^\dagger$ & .82 & .99 & .826 & .86 & .93 & .862 & .85 & .91 & .846 & .98 & .79 & .877 & .98 & .65 & .775 \\
\rowcolor{gray!10}
SSR\,+\,$1/\gamma^\dagger$ (ours) & .87 & .98 & \underline{.869} & .88 & .92 & \underline{.879} & .88 & .86 & \textbf{.852} & .99 & .82 & .895 & .99 & .64 & .770 \\
\rowcolor{gray!10}
SSR\,+\,LS$^\dagger$ (ours) & .91 & .97 & \textbf{.923} & .89 & .90 & \textbf{.885} & .89 & .81 & .848 & .98 & .80 & .882 & .92 & .71 & .799 \\
\midrule
\UB{\textit{Individual erasers}} & \UB{.86} & \UB{.99} & \UB{.919} & \UB{.86} & \UB{.99} & \UB{.918} & \UB{.86} & \UB{.98} & \UB{.916} & \UB{1.00} & \UB{.85} & \UB{.921} & \UB{.99} & \UB{.85} & \UB{.917} \\
\midrule
Leakage $\rho_{\text{single}}$ & \multicolumn{3}{c}{.24} & \multicolumn{3}{c}{.27} & \multicolumn{3}{c}{.26} & \multicolumn{3}{c}{.01} & \multicolumn{3}{c}{.01} \\
Leakage $\rho_{\text{merged}}$ & \multicolumn{3}{c}{.39} & \multicolumn{3}{c}{.56} & \multicolumn{3}{c}{.79} & \multicolumn{3}{c}{.21} & \multicolumn{3}{c}{.51} \\
\bottomrule
\end{tabular}
\caption{Merging $K$ MACE erasers on SD~v1.4 (extends \cref*{tab:style_and_erasure}). E: erasure efficacy ($1{-}\mathrm{Acc}_e$), P: preservation of non-erased concepts ($\mathrm{Acc}_s$), $H_o$: harmonic mean of E, P, and, for CIFAR-10, generality ($1{-}\mathrm{Acc}_g$)~\cite{lu2024mace}; MACE defines no generality term for celebrities. $^\dagger$: uses the preservation term. \textbf{Bold}/\underline{underline}: best/second-best $H_o$. For individual erasers, E is measured per eraser on its own concept, P on the base model, and $H_o$ is the harmonic mean of E and P. Bottom: leakage ratio of the individual erasers and of the merged SSR$^\dagger$ update at $\alpha{=}1$.}
\label{tab:supp:erasure}
\end{table*}

\begin{table}[t]
\centering
\footnotesize
\setlength{\tabcolsep}{4pt}
\begin{tabular}{l l ccc}
\toprule
\textbf{Method} & \textbf{HP} & $K{=}3$ & $K{=}5$ & $K{=}7$ \\
\midrule
Task Arithmetic & $\lambda$ & 0.8 & 0.5 & 0.4 \\
TIES & $\lambda$ & 1.7 & 1.7 & 1.6 \\
DARE & $p$ & 0.3 & 0.5 & 0.1 \\
KnOTS & $\alpha$\,/\,top-$k$ (\%) & 1.0\,/\,20 & 0.8\,/\,40 & 1.0\,/\,20 \\
Core Space & $\alpha$ & 1.0 & 1.0 & 1.0 \\
Iso-C & $\alpha$ & 2.0 & 1.2 & 1.0 \\
SSR\,+\,LS$^\dagger$ (ours) & $\alpha$ & 1.52 & 1.17 & 1.22 \\
\bottomrule
\end{tabular}
\caption{Hyperparameters selected on validation for CIFAR-10 erasure (\cref{tab:supp:erasure}). DARE uses the $\lambda$ of Task Arithmetic; Core Space uses no isotropic scaling. All other methods have no hyperparameters. On celebrities, line search selects $\alpha{=}0.93$ ($K{=}3$) and $0.87$ ($K{=}10$).}
\label{tab:supp:erasure:hp}
\end{table}

\paragraph{When does rescaling help in concept erasure?}

On CIFAR-10, line search selects $\alpha > 1/\gamma$ and even exceeds the individual erasers at $K{=}3$ (\cref{tab:supp:erasure}, left). \Cref{eq:theory:downstream:alpha_star} predicts that this is not universal: $1/\gamma$ is optimal only if the metric prefers the scale of the individual adapters ($s^\star{\approx}1$) and the merge interferes little with inputs it should leave unchanged (small $e$). We test this prediction on celebrity erasure, where both conditions differ.

\emph{Setup.}
We train MACE celebrity erasers~\cite{lu2024mace} for the first 30 of MACE's 100 celebrities and merge $K\in\{3,10\}$ of them (three configurations each). Erasure is scored with MACE's face-identity classifier on its 5 templates $\times$ 5 seeds, preservation on 20 of MACE's other celebrities, and $H_o$ is the harmonic mean of E and P, as in MACE. The preservation term uses 60 further celebrity names with $w{=}3$. Everything else is as above.

\begin{figure}[t]
\centering
\includegraphics[width=0.7\linewidth]{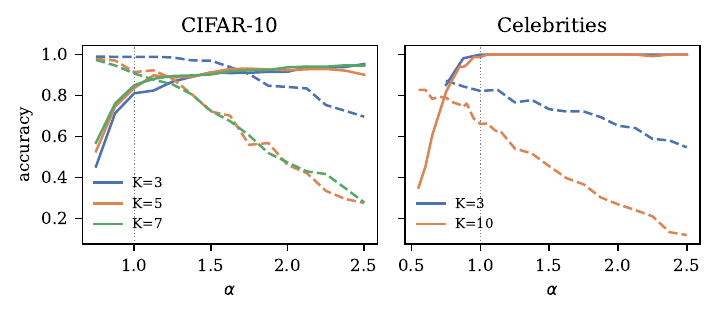}
\caption{Erasure efficacy E (solid) and preservation P (dashed) of SSR$^\dagger$ scaled by $\alpha$, on the validation prompts, averaged over configurations. The dotted line marks the unscaled merge ($\alpha{=}1$). On CIFAR-10 at $K{=}3$, E keeps rising up to $\alpha{\approx}1.5$ while P stays flat. On celebrities, E is already saturated at $\alpha{=}1$ and P drops for any $\alpha{>}1$.}
\label{fig:supp:erasure:alpha}
\end{figure}

\emph{Results.}
On celebrities (\cref{tab:supp:erasure}, right), rescaling helps far less: $1/\gamma$ improves SSR$^\dagger$ at $K{=}3$ (.877 $\to$ .895) but not at $K{=}10$, line search selects $\alpha{<}1$ (0.93 and 0.87), and RegMean$^\dagger$ is best at both $K$. \Cref{fig:supp:erasure:alpha} shows why, with one factor of \cref{eq:theory:downstream:alpha_star} each.

\emph{Preferred scale $s^\star$.} A single MACE object eraser reaches only E\,=\,.86, and scaling the CIFAR-10 merge up raises E until $\alpha{\approx}1.5$, so $s^\star{>}1$. Celebrity erasure is already saturated at $\alpha{=}1$ (E\,=\,.98), so $s^\star{\approx}1$ and larger scales can only cost preservation.

\emph{Interference $e$.} We measure how strongly the merge acts on preserved concepts with a leakage ratio. With the merged SSR$^\dagger$ updates $\Delta W_\ell$ of all key/value projections at $\alpha{=}1$, let $S=\sum_\ell \Delta W_\ell^\top \Delta W_\ell$ and let $\mathcal{E}(c)$ be the mean of $h^\top S h$ over the CLIP text tokens $h$ of the test prompt of concept $c$. Then
\begin{equation}
\rho_{\text{merged}} = \operatorname{mean}_{c\in\text{keep}} \mathcal{E}(c) \,/\, \operatorname{mean}_{c\in\text{erased}} \mathcal{E}(c),
\end{equation}
and $\rho_{\text{single}}$ is the same ratio for the individual erasers. Scaling by $\alpha$ multiplies both energies by $\alpha^2$, but once E saturates only the change on preserved concepts keeps growing. Merging raises leakage, and more so with larger $K$ (bottom of \cref{tab:supp:erasure}). The individual celebrity erasers barely touch other identities ($\rho_{\text{single}}{=}.01$), whereas at $K{=}10$ the merge puts half as much energy on kept identities as on erased ones ($\rho_{\text{merged}}{=}.51$). With $s^\star{\approx}1$ and large $e$, \cref{eq:theory:downstream:alpha_star} places $\alpha^\star$ below $1/\gamma$ and even below 1, which is what line search selects. The same trend appears on CIFAR-10: as $\rho_{\text{merged}}$ grows from .39 ($K{=}3$) to .79 ($K{=}7$), the selected $\alpha$ drops from 1.52 to 1.17--1.22 and the advantage of line search over $1/\gamma$ vanishes (.848 vs.\ .852 at $K{=}7$). Note that $\rho$ alone does not determine $\alpha^\star$: celebrities at $K{=}3$ leak less than CIFAR-10 at $K{=}3$ (.21 vs.\ .39) but have $s^\star{\approx}1$, so their selected $\alpha$ is still lower. $\rho$ is also only a coarse proxy, since it ignores how the denoiser responds to a given change of keys and values.

\emph{Takeaway.}
$1/\gamma$ is a label-free default that helps when erasure is not saturated and the merge leaks little, as on CIFAR-10. When the preferred scale or the leakage changes, line search recovers $\alpha^\star$ in both directions at the cost of a validation sweep.

\end{document}